\documentclass[journal]{IEEEtai}
\usepackage{cite}
\usepackage{amsmath,amssymb,amsfonts}
\usepackage{mdwmath}
\usepackage{array}
\usepackage{graphicx}
\usepackage{multirow}
\usepackage{enumerate}
\usepackage{makecell}
\usepackage{tabularx}
\usepackage{subfigure}
\usepackage{color}

\usepackage[ruled,linesnumbered]{algorithm2e}
\usepackage[caption=false,font=footnotesize,labelfont=sf,textfont=sf]{subfig}

\usepackage{xcolor}
\usepackage{xpatch}
\makeatletter
\def\changeBibColor#1{%
\in@{#1}{}
\ifin@\color{red}\else\normalcolor\fi
}
\xpatchcmd\@bibitem
{\item}
{\changeBibColor{#1}\item}
{}{\fail}
\xpatchcmd\@lbibitem
{\item}
{\changeBibColor{#2}\item}
{}{\fail}
\makeatother

\begin{document}
\title{Efficient Evaluation Methods for Neural Architecture Search: A Survey}

\author{Xiaotian Song, Xiangning Xie, Zeqiong Lv,
       Gary G. Yen,~\IEEEmembership{Fellow,~IEEE},\\
       Weiping Ding,~\IEEEmembership{Senior Member,~IEEE},
       Jiancheng Lv,~\IEEEmembership{Senior Member,~IEEE},
       and Yanan Sun,~\IEEEmembership{Senior Member,~IEEE}
\thanks{Xiaotian Song, Xiangning Xie, Zeqiong Lv, Jiancheng Lv, and Yanan Sun are with the College of Computer Science, Sichuan University, Chengdu 610065, China (e-mails: songxt@stu.scu.edu.cn; xnxie@stu.scu.edu.cn; zq\_lv@stu.scu.edu.cn; lvjiancheng@scu.edu.cn; ysun@scu.edu.cn).\textit{(corresponding author: Yanan Sun.)}}
\thanks{Gary G. Yen is with the School of Electrical and Computer Engineering, Oklahoma State University, Stillwater, OK 74078 USA (e-mail: gyen@okstate.edu).}
\thanks{Weiping Ding is the School of Artificial Intelligence and Computer Science, Nantong University, Nantong, 226019, China, and also the Faculty of Data Science, City University of Macau, Macau 999078, China (e-mail: dwp9988@163.com).}
}

\maketitle
\begin{abstract}
Neural Architecture Search (NAS) has received increasing attention because of its exceptional merits in automating the design of Deep Neural Network (DNN) architectures. However, the performance evaluation process, as a key part of NAS, often requires training a large number of DNNs. This inevitably makes NAS computationally expensive. In past years, many Efficient Evaluation Methods (EEMs) have been proposed to address this critical issue. In this paper, we comprehensively survey these EEMs published up to date, and provide a detailed analysis to motivate the further development of this research direction. Specifically, we divide the existing EEMs into four categories based on the number of DNNs trained for constructing these EEMs. The categorization can reflect the degree of efficiency in principle, which can in turn help quickly grasp the methodological features. In surveying each category, we further discuss the design principles and analyze the strengths and weaknesses to clarify the landscape of existing EEMs, thus making easily understanding the research trends of EEMs. Furthermore, we also discuss the current challenges and issues to identify future research directions in this emerging topic. In summary, this survey provides a convenient overview of EEM for interested users, and they can easily select the proper EEM method for the tasks at hand. In addition, the researchers in the NAS field could continue exploring the future directions suggested in the paper.
\end{abstract}

\begin{IEEEImpStatement}
NAS, a key subfield of AutoML, helps users (e.g., stock analysts and clinicians) easily obtain promising DNN architectures in practice without requiring extensive DNN domain knowledge. However, the performance evaluation process for NAS often proves time-consuming, hindering its widespread application. To address this, many EEMs have been developed to accelerate the performance evaluation process. This paper provides a systematic review of state-of-the-art EEMs, including a novelty categorization method for EEMs, a comprehensive comparative analysis of EEMs, and a summary of the issues in existing EEMs. Moreover, the future research direction and challenges are discussed in detail. By offering a bird's-eye survey of EMMs, this paper facilitates the development of efficient NAS and their broader real-world applications.
\end{IEEEImpStatement}

\begin{IEEEkeywords}
Deep neural network (DNN), neural architecture search (NAS), performance predictors, weight sharing, efficient evaluation
\end{IEEEkeywords}
\IEEEpeerreviewmaketitle

\section{Introduction}\label{Introduction}

\IEEEPARstart{N}{eural} Architecture Search (NAS) aims to automatically discover high-performance Deep Neural Network (DNN) architectures, thus allowing researchers without or with rare expertise to conveniently benefit from the success of DNNs. The architectures designed by NAS have shown to even outperform those manually designed in some tasks~\cite{real2019regularized, zoph2018learning}, and have become increasingly popular in the field of deep learning~\cite{ren2021comprehensive}. Mathematically, NAS formulates the design process of DNN architectures as an optimization problem~\cite{elsken2019neural}. In particular, NAS first defines the search space containing all candidates. Then, it adopts a well-designed search strategy to search for the optimal architecture. During the search process, NAS must evaluate the performance of every searched architecture to effectively guide the running of the search strategy. Generally, the NAS problem is difficult to be solved because of facing multiple optimization challenges, such as the prohibitive computational cost, and with multi-conflicting objectives~\cite{liu2020survey}.

The Evolutionary Computation (EC)~\cite{real2017large,xie2017genetic}, the Reinforcement Learning (RL)~\cite{zoph2018learning,zoph2016neural}, and the gradient algorithm~\cite{fil2021darts} are currently the three mainstream optimization techniques for addressing NAS. The EC-based NAS regards the DNN architectures as the individuals, and iteratively generates the population by applying genetic operators and eliminates poorly performed individuals by selection operators. When the stopping condition is satisfied, the best architecture from the surviving population is picked up for use. In the RL-based NAS, a controller is trained to guide the search process. It uses the performance of the architecture as the reward to update the itself to search for a better architecture in the next iteration. 
The gradient-based NAS generally relaxes the discrete search space to be continuous and uses the gradient descent algorithm to search for a promising architecture. Please note that the gradient-based NAS has been inappropriately claimed to be more efficient than others. This misunderstanding is mainly caused by the representative in this category, i.e., DARTS~\cite{liu2018darts}, which was collectively designed with an Efficient Evaluation Method (EEM) named weight-sharing (will be discussed in Section~\ref{one-shot}). In principle, all NAS algorithms have similar computation complexity if they maintain the same performance evaluation techniques.

Generally, whatever optimization algorithm is used, many DNN architectures need to be evaluated during the search process. This is because these optimization algorithms are iterative, and the current performance must be known in advance, thus effectively guiding the next-step iteration search~\cite{elsken2019neural}. For the Traditional Evaluation Method (TEM), the performance in NAS is evaluated by fully training the corresponding architecture(s) searched in each iteration. Commonly, training a DNN from scratch until converging on a small-scale dataset such as CIFAR-10~\cite{cifar10dataset} may take hours or even days depending on the scale of the DNN. Consequently, since there are often thousands of DNNs to be trained in NAS, the whole NAS algorithm becomes prohibitively computation-intensive and time-consuming. For example, on CIFAR-10, the LargeEvo algorithm~\cite{real2017large} consumed 250 Graphic Processing Units (GPUs) for 11 days. The NAS-RL algorithm~\cite{zoph2016neural} took 800 GPUs for 28 days. Even more, the RegularizedEvo algorithm~\cite{real2019regularized} ran on 450 GPUs for 7 days. In practice, buying or renting such scales of GPU resources is commonly unaffordable for most researchers~\cite{xie2022benchenas}. As a result, how to accelerate the TEM to reduce the prohibitive computational overhead is essential, which results in the research topic of EEMs in the NAS community~\cite{ren2021comprehensive}. Specifically, EEM refers to the performance evaluation method that consumes less time than TEM in the entire performance evaluation process of NAS.

\begin{figure}[!t]
\centerline{\includegraphics[width=0.43\textwidth]{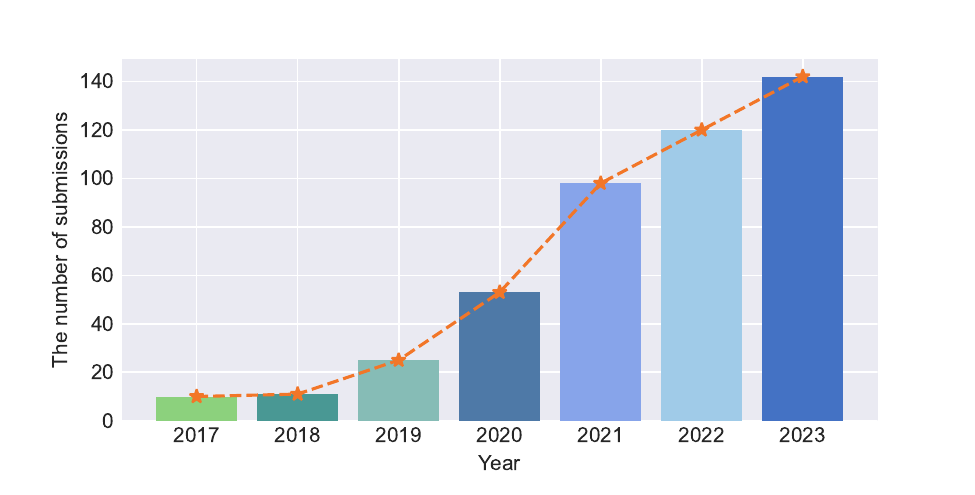}}
  \caption{The number of ``submissions'' refers to EEMs. The statistical results are searched on Google Scholar with the following steps: (1) Select a specific year; (2) Search with the keywords ``early stopping'' OR ``learning curve'' OR ``network morphism'' OR ``weight sharing'' OR ``one-shot'' OR ``DARTS'' OR ``differentiable'' OR ``predictor'' OR ``population memory'' OR ``zero-cost'' OR ``zero-shot'' OR ``training-free'' AND ``architecture search'' OR ``architecture design'' OR ``CNN'' OR ``deep learning'' OR ``deep neural network''; (3) Check the selected papers in detail to verify if they belong to EEMs.}\label{fig_submission}
\end{figure}

To the best of our knowledge, the first work of EEMs is designed in the LargeEvo algorithm~\cite{real2017large} preprintly available in Arxiv in 2016, although there are also some earlier works that potentially can achieve the same goal~\cite{domhan2015speeding}. With the development of NAS, the research of EEMs has received much attention and is becoming one of the hottest topics in the current artificial intelligence community. As evidenced by Fig.~\ref{fig_submission} which reports the submissions of EEMs from 2017 to 2023, the number of submissions increases year by year. Despite the popularity and criticality of the EEMs, there is a lack of a survey to systematically review these works. This will make it difficult for researchers with rarely relative knowledge to quickly grasp the current situation and landmark works in the research topic.

Related surveys~\cite{ren2021comprehensive, liu2020survey, elsken2019neural, baymurzina2022review} primarily focus on the development of NAS, with little discussion on EEMs. However, the efficient evaluation is a crucial component of NAS methods, and a rough number of 300 publications, as we have reported above, has been devoted to EEMs for efficient NAS during past years. To the best of our knowledge, only Liu \textit{et al.}~\cite{liu2022survey} provided an overview of efficient NAS. However, their survey lacks systematic research on EEMs. First, they miss some important EEMs such as the zero-shot methods discussed in Subsection~\ref{zero-shot}, and do not discuss the evaluation criteria of EEMs. Second, their categorization method only considers the consumed time of EEMs during the running of NAS, but ignores the pre-computation time which is also crucial for constructing the EEMs. In this survey, we aim to systematically encompass all existing EEMs and subsequently classify them into four distinct categories, employing a novel categorization method to provide readers with a clearer and more effective way of comprehending the variety of available EEMs. In summary, the contributions of our paper are as follows:
\begin{itemize}
\item This survey provides a comprehensive and systematic discussion of EEMs. With this survey, the interested researchers could be easily informed about the development and the taxonomy of EMMs which is one of the hottest research topics in the community of artificial intelligence.
\item We have conducted a thorough analysis to summarize the advantages and disadvantages of various EEMs through an exhaustive investigation of close to 200 references in recent years. This work can provide valuable insights to NAS researchers, facilitating the swift selection of appropriate EEMs and the development of efficient NAS.
\item We analyze the current status and future development directions of each specific EEM, pointing out the challenges that need to be addressed in the future of the EEMs field. This will serve as a guideline for researchers to investigate impactful EEMs and promote the development of automated machine learning.
\end{itemize}

In the following, we show the detail of the categorization and organization in Section~\ref{org}. After that, the details of these EEMs are surveyed in Sections~\ref{section_method} to~\ref{direction}, including the method designs, evaluation metrics, and future directions. Finally, we make a conclusion in Section~\ref{conclusion}.

\section{Categorization and Organization}\label{org}
In this section, we first give the formulation of NAS in Subsection~\ref{formula_NAS}. Then, we describe the notation and terminology used in this paper in Subsection~\ref{org_pd}. After that, the categorization for the EEMs is detailed in Subsection~\ref{cate}. Finally, the organization of this survey is discussed in Section~\ref{section_org}.

\subsection{Formulation of NAS}\label{formula_NAS}

As discussed in Section~\ref{Introduction}, the optimization problem of NAS can be formulated as Equation~(\ref{NAS_equ}):
\begin{equation}\label{NAS_equ}
\mathop{\arg\max}\limits_{A\in\mathcal{A}} \mathcal{L}(A, \mathcal{D}_{\text{train}}, \mathcal{D}_{\text{valid}})
\end{equation}
where $\mathcal{A}$ represents the search space that filled with a set of neural architectures, $\mathcal{L}(\cdot)$ denotes the performance of neural architecture $A$ on the validation dataset $\mathcal{D}_{\rm{valid}}$ after being trained on the training dataset $\mathcal{D}_{\rm{train}}$. As for the DNNs on image classification tasks, for example, $\mathcal{L}(\cdot)$ denotes the classification accuracy.

\begin{figure}[!t]
\centerline{\includegraphics[width=0.40\textwidth]{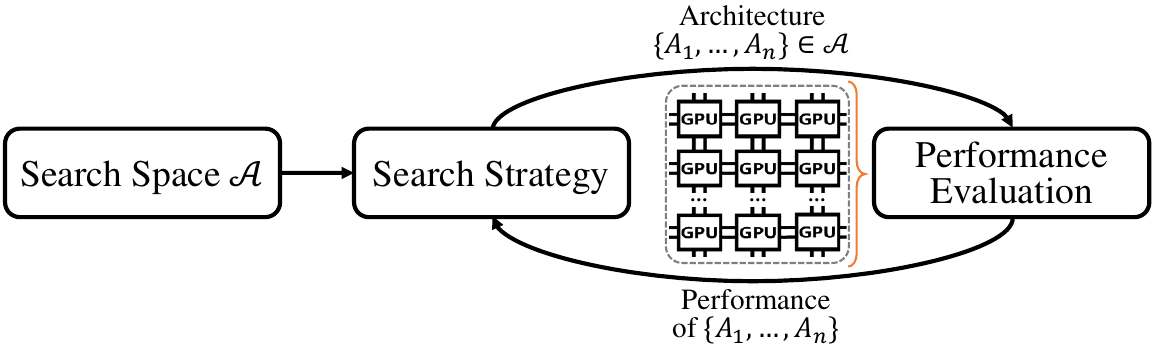}}
	\caption{The workflow of the NAS algorithm.}\label{fig_NAS_process_TEM}
\end{figure}

The optimization problem in Equation~(\ref{NAS_equ}) can be solved with three stages, as shown in Fig.~\ref{fig_NAS_process_TEM}: search space, search strategy, and performance evaluation. Specifically, the search process occurs in the predefined search space, and the search strategy is used to select the promising architecture in the search space. However, the architectures selected by the search strategy are typically unlabeled, i.e., the performance of the architectures is unknown. Thus, we cannot verify whether the architecture is good or not. Performance evaluation aims to address this, which can get the $\mathcal{L}(\cdot)$ for each architecture to guide the search process. Typically, the search-evaluation process iterates in a loop and needs to evaluate numerous architectures (more than hundreds). This process for the TEMs requires evaluating each architecture on the GPUs. It is computationally extensive and time-consuming. EEMs, therefore, try to reduce and even eliminate the requirements of the GPUs, which can significantly accelerate the NAS algorithm.

\subsection{Notation and Terminology}\label{org_pd}

As introduced in Subsection~\ref{formula_NAS}, NAS aims to find the architecture $A$ with promising performance. Thus, the evaluation process of NAS needs to get the performance of every searched architecture, as indicated in Equation~(\ref{NAS_equ}). Note that we use $\mathcal{A}^{s}$ to denote the architectures searched in all iterations of the NAS algorithm.

Based on the time consumption of the evaluation process, existing evaluation methods can be divided into TEM and EEM. To better describe them, we first define the \emph{\textbf{initialization time}}, \emph{\textbf{evaluation time}}, and \emph{\textbf{runtime}}. Specifically, the initialization time is the time used to evaluate the architecture on GPUs to pre-build the evaluation method, while the evaluation time refers to the overall time for directly evaluating the performance of the architecture on GPUs. The runtime is the sum of the initialization time and the evaluation time, which can be represented as Equation~(\ref{euqation_eval}):
\begin{equation}\label{euqation_eval}
T = \mathcal{T}_{train}(\mathcal{A}^{in}) + \mathcal{T}^{*}_{train}(\mathcal{A}^{ev})
\end{equation}
where $\mathcal{T}_{train}(\cdot)$ represents the time for training the architectures and building the evaluation method, and $\mathcal{T}^{*}_{train}(\cdot)$ denotes the time to only evaluate architectures on GPUs. In addition, $\mathcal{A}^{in}$ represents the architectures used to build the evaluation method, and $\mathcal{A}^{ev}$ denotes architectures that need to directly obtain their performance on GPUs.

\textbf{TEM} needs to fully train each searched architecture on GPUs, and its initialization time is zero since there is no pre-computation. Specifically, $\mathcal{A}^{TEM}$ represents the architectures that require evaluation of TEM, which is equal to $\mathcal{A}^{s}$. Thus, the runtime of TEM is $T_{TEM}=\mathcal{T}^{*}_{train}(\mathcal{A}^{ev})$, subject to $\mathcal{A}^{TEM} = \mathcal{A}^{ev}$. This indicates that each searched architecture requires training, resulting in TEM being time-consuming.

\textbf{EEM} consumes less runtime than TEM for the same NAS algorithm. It is achieved through different working principles. One is directly reducing the evaluation time with training acceleration methods. The other part involves evaluating the performance of few searched architectures and building evaluation methods based on them, which takes initialization time. In particular, $\mathcal{A}^{EEM}$ represents the architectures utilized in EEM. Therefore, the runtime of EEM can be represented as $T_{EEM}=\mathcal{T}_{train}(\mathcal{A}^{in}) + \mathcal{T}^{*}_{train}(\mathcal{A}^{ev})$, subject to $\mathcal{A}^{EEM} = \mathcal{A}^{in} + \mathcal{A}^{ev}$.  Note that $\mathcal{A}^{EEM}$ differs from $\mathcal{A}^{s}$ and different EEMs can be categorized based on the different $\mathcal{A}^{EEM}$.

\subsection{Categorization}\label{cate}
Existing categorization methods have some limitations. Specifically, White \textit{et al.}~\cite{white2021powerful} divided existing EEMs into four categories: model-based, learning curve-based, zero-cost proxies, and weight sharing. However, they categorize the existing EEMs based on their working principles. This means once an EEM with a new working principle appears, this categorization method may no longer apply. In fact, the downscaled datasets methods and the downscaled model methods mentioned in Section~\ref{$N$-shot} do not belong to the four categories. Liu \textit{et al.}~\cite{liu2022survey} proposed to categorize existing EEMs into proxy-based and surrogate-assisted methods. The proxy-based methods need the weights of the neural network in the evaluation process, while the surrogate-assisted methods are just the opposite. The reason for this categorization is that the proxy-based methods require additional computing resources during the running of the NAS algorithm, while the surrogate-assisted methods do not. Although this method can stably classify EEMs, it only considers the evaluation time and ignores the initialization time which is also crucial. As a result, existing categorization methods cannot accurately measure the true time complexity of EEMs.

In this survey, we categorize the existing EEMs based on $|\mathcal{A}^{EEM}|$. 
The categorization method in this survey is inspired by the concept of ``$K$-shot learning’’~\cite{wang2020generalizing} from the field of machine learning. Specifically, ``$K$-shot learning’’ refers to a machine learning paradigm where the model is trained with ``$K$'' annotated samples per class. This term serves as a generalization that encompasses various research topics, including zero-shot learning, one-shot learning, and few-shot learning. Considering this concept has been accepted a convention in the machine learning community, in this survey, we also adopt the term ``shot'' to describe the number of annotated DNNs during the NAS process and categorize EEMs into zero-shot, one-shot, few-shot, and $N$-shot methods. In general, the more annotated samples used in the training process, the more complex of the corresponding machine learning algorithm will be. In this regard, this categorization method can also intuitively infer the time complexity of the target EEMs, which in turn helps the corresponding researchers easily choose the EEMs with the proper complexity for the use. Specifically, this results in four different categories shown below:
\begin{itemize}
\item \emph{$N$-shot evaluation method}. For the $N$-shot evaluation method, the number of trained architectures is greater than or equal to the number of searched architectures, i.e., $|\mathcal{A}^{EEM}|\geq|\mathcal{A}^{s}|$. The $N$-shot evaluation methods still need to train every searched architecture, and mainly accelerate the training of these architectures to consume less runtime than TEM.

\item \emph{Few-shot evaluation method}. For the few-shot evaluation method, the number of trained architectures is less than the number of searched architectures and greater than one, i.e., $|\mathcal{A}^s|>|\mathcal{A}^{EEM}|>1$. Because the number of trained architectures of the few-shot evaluation method is less than the number of searched architectures, the runtime is naturally less than $T_{TEM}$. 

\item \emph{One-shot evaluation method}. For the one-shot evaluation method, the number of trained architectures is equal to one, i.e., $|\mathcal{A}^{EEM}|=1$. The one-shot evaluation method can consume less runtime than TEM because only one architecture needs to be trained.

\item \emph{Zero-shot evaluation method}. For the zero-shot evaluation method, the number of trained architectures is equal to zero, i.e., $\mathcal{A}^{EEM}=\emptyset$. This method involves no training, thus resulting in an extremely low cost.
\end{itemize}

We hereby acknowledge that the proposed categorization method has certain limitations and potential bias. In particular, the efficiency of EEMs is not solely determined by the time spent training DNNs in $\mathcal{A}^{EEM}$. For zero-shot methods where training neural networks is unnecessary, the process still requires feeding one or more batches of data into each network for multiple forward and backward passes, which consumes time. If, in future EEMs, non-training processes account for the majority of runtime, the interpretability of this categorization method could be compromised.

\subsection{Organization}\label{section_org}
\begin{figure*}[htbp]
	\centerline{\includegraphics[width=0.83\textwidth]{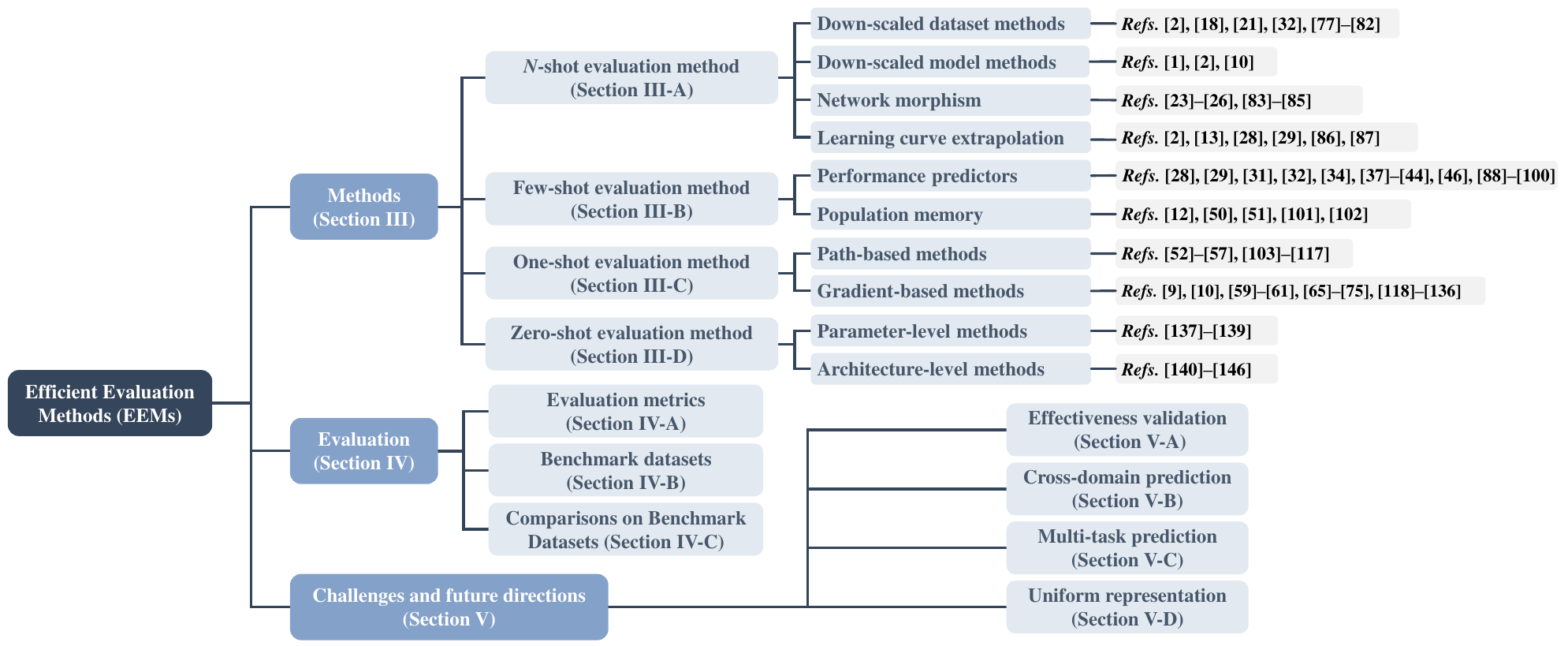}}
	\caption{The organization of the description for EEMs.}
	\label{categorization}
\end{figure*}

The $N$-shot evaluation methods, the few-shot evaluation methods, the one-shot evaluation methods, and the zero-shot evaluation methods are discussed in Section~\ref{section_method}, with the focus on design principle, strength and weakness analysis of these methods. In Section~\ref{evaluation}, we detail the evaluation for these EEMs, including the evaluation metrics, benchmark dataset, and the comparison results on benchmark datasets. In Section~\ref{direction}, we discuss the challenges and future directions. For the convenience of quickly navigating to the interested part, an illustration of above organization is shown in Fig.~\ref{categorization}.

\section{Methods}\label{section_method}

In this section, we discuss the working principle and research status of various EEMs under the four categories. For each method, we first give its main operations or workflow. Then, detailed literature examples are provided. Finally, we discuss the advantages and disadvantages of the method.

\subsection{$N$-shot Evaluation Methods}\label{$N$-shot}
\begin{figure}[htbp]
\centerline{\includegraphics[width=0.38\textwidth]{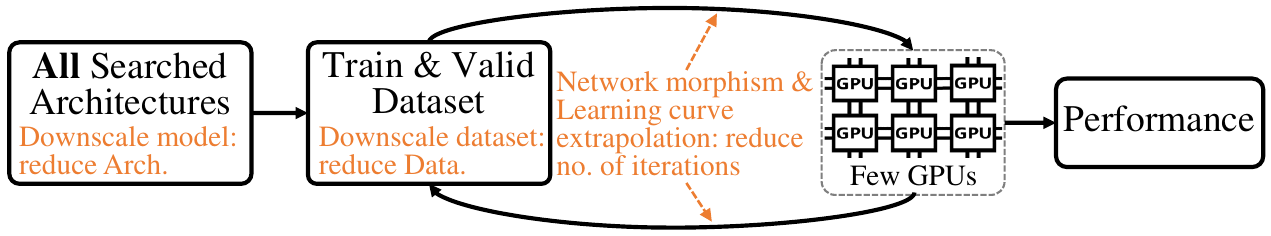}}
	\caption{The flowchart of $N$-shot evaluation method and its relationship to subcategory methods.}\label{N_shot_workflow}
\end{figure}

As shown in Fig.~\ref{N_shot_workflow}, for all searched architectures, the $N$-shot evaluation methods still need to be trained and validated on the dataset using GPUs. Different subcategories of $N$-shot methods functioned at different stages to accelerate the evaluation process.

For the purpose of better discussion, we first describe the training process for a DNN. During the training process, all samples in the dataset are sent into the model (i.e., architecture) to complete one forward and backpropagation, and the process repeats $N_{epoch}$ times. The training time mainly depends on three factors: the dataset scale, the model size, and the number of epochs. Specifically, the time of completing one forward and backpropagation depends on the dataset scale and the model size. If the dataset scale or the model size is large, the computational complexity becomes higher, which leads to much time spent on the computation. Furthermore, the training time is also positively correlated with the number of epochs ($N_{epoch}$). The larger $N_{epoch}$, the longer the training time.

The $N$-shot evaluation methods mainly consist of \emph{downscaled dataset methods}, \emph{downscaled model methods}, \emph{network morphism}, and \emph{learning curve extrapolation}. Please note that $\mathcal{A}^{EEM}$ is equal to $\mathcal{A}^s$ (i.e., $\mathcal{A}^{EEM}=\mathcal{A}^s$) for all $N$-shot evaluation methods except for the learning curve extrapolation. As the name says, the downscaled dataset methods aim to reduce the dataset scale to accelerate the training. The downscaled model methods focus on reducing the model size. Furthermore, network morphism and learning curve extrapolation aim to reduce the number of epochs.

\subsubsection{Downscaled Dataset Methods}
Similar to TEM, the downscaled dataset methods train every searched architecture to obtain their performance to guide the search process. The difference is that they use a downscaled dataset to replace the original one to accelerate the training of architectures. Naturally, the runtime of the downscaled dataset methods is less than that of TEM. The downscaled dataset methods can be divided into three categories: (1) samples a subset of the original dataset; (2) downsamples each sample in the original dataset; and (3) uses a different and smaller proxy dataset with similar properties to the original dataset. All of them do not rely on pre-build evaluation methods, and the runtime of them only contains evaluation time. Please note that the ``proxy data'', ``proxy dataset'', and ``data proxy'' are the keywords to select specific methods. 

\textbf{Category (1): }This category samples a subset of the original dataset as the downscaled dataset. Because the number of samples in the dataset is reduced, the overall dataset scale is reduced. The simplest way to obtain the subset is to randomly select some samples from the original dataset. \textit{\textbf{Literature examples:}}  Liu \textit{et al.}~\cite{liu2019deep} randomly selected a small number of medical images from the entire dataset to train the searched architectures. However, random sampling may lead to the removal of some representative samples, thus greatly weakening the generalization of the architecture trained in the subset~\cite{park2019data, na2021accelerating}. This will lead to an inaccurate evaluation of the performance. To alleviate the problem, some works designed new sampling methods. For example, Park \textit{et al.}~\cite{park2019data} designed a probe network to measure the impact of every sample on the performance ranking of the architectures. Then, they removed the samples that have a small impact on the performance of architectures. Na \textit{et al.}~\cite{na2021accelerating} used the data entropy to analyze existing sampling methods, and proposed a new sampling method that prefers samples in the tail ends of the data entropy distribution.

\textbf{Category (2):} This category downsamples each sample in the original dataset to reduce the dataset scale. Because the size of every sample becomes smaller by the downsampling, the overall dataset scale naturally becomes smaller. A typical way is to reduce the resolution of the images in the dataset for image classification. Specifically, as the resolution of the image is reduced, the pixels in the image become fewer, and the size of the image naturally becomes smaller. As a result, reducing the image resolution in a dataset can reduce the dataset scale, thus speeding up the training.  \textit{\textbf{Literature examples:}} Chrabaszcz \textit{et al.}~\cite{chrabaszcz2017downsampled} proposed the variants of ImageNet~\cite{deng2009imagenet}, i.e., ImageNet16x16, ImageNet32x32, and ImageNet64x64. Specifically, ImageNet16x16, ImageNet32x32, and ImageNet64x64 used the box technique to downsample all images in ImageNet to 16x16, 32x32, and 64x64, respectively.

\textbf{Category (3):} This category uses a different and smaller proxy dataset with similar properties to the original dataset to train the searched architectures. The proxy dataset may have fewer samples and/or smaller sample size compared with the original dataset. \textit{\textbf{Literature examples:}} Zoph \textit{et al.}~\cite{zoph2018learning} used CIFAR-10 as the proxy for ImageNet because CIFAR-10 not only has both fewer samples and a smaller size than ImageNet but also targets the same image classification task as ImageNet.

{\textbf{Advantages:}} The downscaled dataset method is straightforward and easy to implement. It can reduce one to several orders of magnitude computational costs depending on the size of the downscaled dataset and the original dataset.

\textbf{Disadvantages:} In practice, the performance rankings of DNNs trained on the original dataset may not be consistent with their performance rankings when trained on a downscaled dataset. This inconsistency is caused by two main factors. (1) Information Loss: The downscaled datasets will inevitably lose some information present in the original dataset. The lost information may contain important features and patterns that impact the performance. Therefore, the abilities of certain models that perform well on the original dataset may not be adequately reflected on the downscaled dataset. (2) Class Imbalance: Downscaling can introduce the imbalance proportions of the samples in various classes. If the relative proportions of classes are not preserved during downscaling, certain class samples may be excessively reduced, affecting the learning ability from those classes. This can lead to performance degradation, particularly in minority classes, when trained on the downscaled dataset, deviating from the performance on the original dataset. In order to avoid these problems as much as possible, the downscaled methods should be carefully designed to maintain class balance and to avoid the loss of information with a significant impact on performance.

\subsubsection{Downscaled Model Methods}\label{subsec:downscaled} The downscaled model methods reduce the model size during the search process (e.g., search one cell rather than the whole architecture), and the optimal cell found is repeated after the search process. Similar to downscaled dataset methods, the runtime of the downscaled model methods only needs evaluation time.

\textbf{Literature examples:} Many works have used this method to accelerate the performance evaluation process, and we use the keywords ``cell repeat'' and ``repeated module'' to choose the papers mentioned below. For example, Zoph \textit{et al.}~\cite{zoph2018learning} proposed a cell-based search space (i.e., NASNet search space) which repeatedly stacked the same cell structure to build the architecture, and only the cell structure needs to be found during the search process. To reduce the model size, they transformed each model to one with a fewer number of cell repeats and filters in the initial convolutional cell during the search. Because of its efficiency, many works~\cite{real2019regularized, liu2018darts} used the NASNet search space or its variants and followed its downscaled model method to accelerate the training. 

\textbf{Advantages:} The downscaled model method is also easy to implement, and it can save GPU memory. In addition, the training time becomes less and the runtime spent on performance evaluation is less than $T_{TEM}$.

\textbf{Disadvantages:} Downscaling methods usually assume that the same downscaling action has the same effect on the performance of different models. This is because the performance ranking of the downscaled model and the original model will remain consistent only if the degree of the performance change is the same for all models. However, this assumption does not always hold. For example, removing a convolution operation from a high-complexity model may improve its performance because the operation may be redundant. Removing the same convolution operation from a low-complexity model may reduce its performance because of the removal of the learnable operation. As a result, these methods may result in a lower correlation between the performance of the downscaled model and that of the original model. In the future, this method should theoretically demonstrate the correlation between the performance ranking of the original model and the downscaled model. This will help in the development of a more rigorous downscaled model method, thereby avoiding the sole reliance on intuition when downsizing the original model.

\subsubsection{Network Morphism} Network morphism can accelerate the evaluation process of most searched architectures by reducing the number of epochs, so the evaluation time is the only cost of the runtime of network morphism. Please note that not every training process of the searched architecture is accelerated, the reason for which is explained in the third paragraph of this subsection. Because the training time for most searched architectures is reduced, the runtime of network morphism is significantly less than that of TEM. Network morphism is first proposed in the background of transfer learning to rapidly transfer knowledge from one fully-trained network (i.e., a parent network) into another network (i.e., a child network). It is soon applied in the field of NAS because it can effectively accelerate the training of the child network. For the sake of understanding, we first explain the acceleration principle of network morphism and then discuss how it is applied in NAS as an EEM.

\begin{figure}[!t]
\centerline{\includegraphics[width=0.32\textwidth]{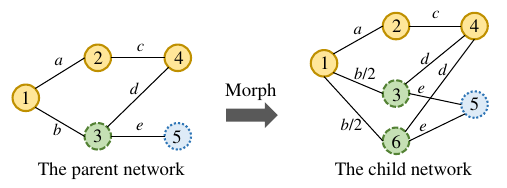}}
  \caption{An example of width morphing, where different types of nodes indicate distinct operations (e.g., convolution layer, pooling layer). The labels on the edges represent the corresponding weight values. The parent network is a fully-trained network, and the child network is morphed from the parent network by a width morphing operation. Specifically, in the child network, node \#6 is a copy of node \#3, and the weights of node \#3 are also transferred to node \#6. To preserve the function of the two networks, the value of weight $b$ is divided by two, while the values of weights $a$, $c$, $d$, and $e$ remain unchanged. In this way, if the same input is fed to the parent network and the child network, respectively, the output of both networks will be the same, which means the function of the parent network is preserved in the child network.}\label{fig_network_morphism}
\end{figure}

The term \emph{morphism} mathematically means a structure-preserving map from one structure to another of the same type. Network morphism refers to a function-preserving action to transform a fully-trained parent network into a child one that can completely preserve the function of the parent network~\cite{wei2016network}. Specifically, the parent network is first transformed into a different child network by predefined morphing actions. Then, the child network directly inherits the weights from the parent network and has the same function and output as the parent network. After that, the child network is generally trained on the dataset. Because the child network does not require being trained from scratch, the number of epochs needed to train to convergence is naturally reduced. As the inheriting process does not need intensive resources, the network morphism can accelerate the training time of the child network. The morphing actions (i.e., function-preserving actions) mainly include width morphing, deep morphing, kernel size morphing, etc. Different types of morphing change different parts of the parent network to generate a new child network. For the convenience of understanding, Fig.~\ref{fig_network_morphism} shows an example of width morphing. As the name says, the child network is wider than the parent network. However, the functions performed by both the child network and the parent network remain the same.

Generally, network morphism cannot be used to accelerate the evaluation in NAS because there must exist fully-trained parent networks for using network morphism. To gap this, the architectures searched in the first iteration of the search strategy are generally fully trained as the parent of the architectures searched in the next iteration. Specifically, the architectures searched in the first iteration of NAS are fully trained as the parent network. Then, various morphing operations are applied to these parent networks to generate child networks. These child networks inherit weights from the parent network and are trained to obtain their performance with fewer epochs. These child networks are then regarded as the parent network, and the above process is iteratively performed. Because all searched architectures except the ones searched in the first iteration are trained with fewer epochs compared with TEM, network morphism consumes less runtime than $T_{TEM}$.

\textbf{Literature examples:} 
Many works have used network morphism to accelerate the performance evaluation process, and the keywords of them are ``network morphism'' and ``network transformation''.
For example, Cai \textit{et al.}~\cite{cai2018efficient} adopted the RL-based search strategy, and used the controller to generate a morphing action to apply in the current network for the search. The designed morphing actions include a width morphing called Net2Wider and a deep morphing called Net2Deeper. Net2Wider can replace a layer with a wider layer such as more filters for convolutional layers, and Net2Deeper inserts a new layer that is initialized as an identity mapping between two layers. Then, the newly searched architectures inherited weights from the current network and were trained with fewer epochs to obtain their performance.
Cai \textit{et al.}~\cite{cai2018path} proposed the path-level morphing actions because the previous work adopted layer-level morphing which can only add filters or add layers. The path-level morphing can modify the path topologies of the parent networks and preserve their functions. Noting that the previous morphing actions can only increase the size of the networks because the function-preserving property is not guaranteed when decreasing the size of the networks. 
This is not suitable for some computationally constrained scenarios. 
To solve the problem, Elsken \textit{et al.}~\cite{elsken2018efficient} designed the approximate network morphism that could decrease the network size and roughly preserve the functions. Then, they adopted the EC-based search strategy and used the approximate network morphism as a mutation operator to search for architectures.

\begin{figure}[!t]
\centerline{\includegraphics[width=0.39\textwidth]{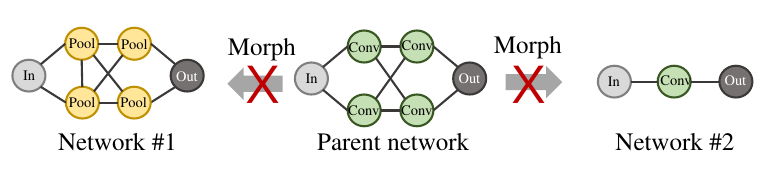}}
  \caption{An example to show the limit of the network morphism. Network \#1, Network \#2 and the parent network are from the same search space, and the parent network is a fully-trained network. Network \#1 and Network \#2 cannot be morphed from the parent network. Specifically, Network \#1 cannot be morphed from a parent network with different operations. This is because the operation types (except for input and output nodes) are different between the parent network and Network \#1. Network morphism only modifies the topological architecture of neural networks without changing the operation types of existing neurons. Network \#2 cannot be morphed from the parent network because the function of the parent network with four convolution layers cannot be the same as Network \#2 with only one convolution layer. }\label{fig_network_morphism_limit}
\end{figure}

\textbf{Advantages:} The network morphism method can leverage much knowledge from parent networks, and it has more theoretical foundations.

\textbf{Disadvantages:} 
Not all the networks in the search space can be transformed from a fully-trained parent network. In this case, the network morphism method will fail. Specifically, the possible neural architectures morphed by the existing morphism operation (e.g., depth morphing, width morphing, kernel size morphing, and subnet morphing) are limited. As shown in Fig.~\ref{fig_network_morphism_limit}, not all potential architectures can be explored during the search process, thereby restricting the diversity and innovation of searched architectures. In addition, network morphism may lead to the search process getting trapped in local optima. This is because the global optimal may not be obtained by morphing the parent network, which causes the algorithm to fall into a local optimal. Specifically, assuming that Network \#2 in Fig.~\ref{fig_network_morphism_limit} represents a globally optimal solution and the parent network is the current solution, it becomes evident that relying solely on morphing operations cannot reach the ultimate global optimum. Such limitations can restrict the exploration of the search space and prevent NAS from finding the optimal network architecture. Overall, the limitations of network morphism stem from the lack of diversity of existing morphing operations. One future direction is to explore more types of morphing operations to promote the diversity of morphed networks. In this way, the inflexibility of network morphism can be alleviated to some extent.

\subsubsection{Learning Curve Extrapolation}\label{learning_curve}
Different from other $N$-shot evaluation methods that $\mathcal{A}^{EEM}$ is equal to $\mathcal{A}^s$, $\mathcal{A}^{EEM}$ in the learning curve extrapolation methods composed of $\mathcal{A}^s$ and $\mathcal{A}^{f}$, i.e., $\mathcal{A}^{EEM}=\mathcal{A}^s\cup \mathcal{A}^{f}$, where $\mathcal{A}^{f}$ is an architecture set that includes some fully trained architectures for learning curve extrapolation. The learning curve extrapolation trains a model on $\mathcal{A}^{f}$ to predict the actual performance of the architecture trained after only a few epochs. In this way, the training time of the searched architectures is largely reduced because the epochs become fewer. Since the time saved by training the architectures in $\mathcal{A}^s$ is generally greater than the time spent on fully training the architectures in $\mathcal{A}^f$, it can still spend less time on performance evaluation than TEM.

To illustrate the principle of learning curve extrapolation, we first introduce three terms: \emph{learning curve}, \emph{partial learning curve}, and \emph{final learning curve}. Specifically, the term learning curve generally refers to the function of performance with a growing number of iterations for an iterative machine learning algorithm. We use the term final learning curve to represent the entire learning curve $f_{t}=(p_{1},p_{2},\dots,p_{t}$) of an algorithm from the beginning to the end of the training, where $p_{i}$ represents the performance at the iteration of $i$ and $p_{t}$ is the final performance. Furthermore, we use the term partial learning curve to refer to the learning curve observed as of epoch $l$, i.e., $f_{l}=(p_{1},p_{2},\dots,p_{l}) (l<t)$.

The workflow of the learning curve extrapolation method is as the following. First, some architectures are sampled from the predefined search space, and then are fully trained to construct the architecture set $\mathcal{A}^f=\{(x^{1},f_{t}^{1}),(x^{2},f_{t}^{2}),\dots,(x^{k},f_{t}^{k})\}$ where $x^{i}$ denotes the $i$-th architecture and $f_{t}^{i}$ represents its final learning curve. Then, the architecture set is transformed to another form $\mathcal{A}^f=\{(x^{1},f_{l}^{1},p_{t}^{1}),(x^{2},f_{l}^{2},p_{t}^{2}),\dots,(x^{k},f_{l}^{k},p_{t}^{k})\}$ where $f_{l}^{i}$ represents the partial learning curve and $p_{t}^{i}$ denotes the final performance of architecture $x^i$. The learning curve extrapolation method builds a model $\mathcal{P'}(x^{i},f_{l}^{i})$ by training it on $\mathcal{A}^f$. The model $\mathcal{P'}(\cdot)$ can be used to predict the final performance of the searched architectures by feeding their architectures and partial learning curves. 
Finally, the model is used to predict the final performance of the searched architectures after obtaining their partial learning curve. {The runtime of the learning curve extrapolation method includes initialization time (to build the predicted model) and evaluation time (to get the partial learning curve).} 

It is worth mentioning that the early-stopping strategy, a popular EEM, is considered a special case of learning curve extrapolation. Specifically, the early-stopping strategy directly regarded the obtained performance after just a few epochs as the final performance. We can view that the early-stopping strategy builds a function $p_{t}=f_{l}(t)=p_{l}$ for a partial learning curve $f_{l}=(p_1,p_2,\dots,p_l)$ ($l< t$) to predict the final performance $p_{t}$. Because no training architecture is needed to build $p_{t}$, $\mathcal{A}^{EEM}$ is equal to $\mathcal{A}^s$ for the early-stopping strategy.

\textbf{Literature examples:} 
The keywords we used to select the studies of the learning curve extrapolation are ``partial learning curves,'' ``partial training,'' and ``early stop''. The learning curve extrapolation was first proposed to automatically find a well-performing hyperparameter configuration for neural networks~\cite{rijn2015fast}. Because both hyperparameter optimization and NAS are faced with the problem of the high cost of network evaluation, the learning curve extrapolation is soon applied to NAS.
For example, Rawal \textit{et al.}~\cite{rawal2018nodes} developed a Long Short Term Memory (LSTM) consisting of an encoder RNN and a decoder RNN to predict the final performance of a partially trained model. They fed the validation loss at the first $10$ epochs to the encoder and obtained the final validation loss by the decoder. Baker \textit{et al.}~\cite{baker2017accelerating} used a set of support vector machines to predict the architecture performance by feeding the features of model architectures, training hyper-parameters, and partial learning curve. Furthermore, they model the estimation process as a gaussian perturbation to calculate the probability $p(\hat{y}\leq y_{best})$ where $\hat{y}$ represents the estimated final performance, $y_{best}$ denotes the performance of the best architecture. When $p(\hat{y}\leq y_{best})\geq \Delta$ where $\Delta$ is used to balance the trade-off between increased speedups and risk of prematurely terminating good architecture, the training of an architecture terminates. Wistuba \textit{et al.}~\cite{wistuba2020learning} built a neural network to rank the architecture with the input of architecture, partial learning curve, dataset, and hyper-parameters. In this way, they can search for the optimal architecture. As for the early-stopping strategy, it can be easily embedded in the NAS algorithms. For example, Zoph \textit{et al.}~\cite{zoph2018learning} trained every architecture for only 20 epochs to reduce the search cost.

\textbf{Advantages:} The learning curve extrapolation can decrease the number of epochs to accelerate the training process. It is easy to combine with different NAS algorithms without modifying data or models. Thus, it is highly flexible.

\begin{figure}[!t]
  \centerline{\includegraphics[width=0.25\textwidth]{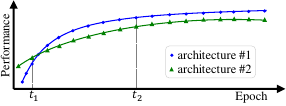}}
  \caption{An example of the early stopping strategy causing inaccurate prediction. The figure shows the performance values of architecture \#1 and architecture \#2 at every epoch. The performance ranking at $t_1$ of architecture \#1 and architecture \#2 is not consistent with the final performance ranking at $t_2$. This is because architecture \#2 is close to convergence at $t_1$, while architecture \#1 still requires further training. Hence, when both architectures converge, architecture \#1 has higher performance than architecture \#2.}\label{fig_learning_curve}
\end{figure}

\textbf{Disadvantages:} 
The learning curve extrapolation needs to fully train some architectures as the dataset for building the prediction models. The amount of data should not be too small, otherwise, the generalization ability of the prediction model may be poor. Thus, it still requires consuming much time to collect the dataset. For example, Baker \textit{et al.}~\cite{baker2017accelerating} built three architecture datasets for ResNets, MetaQNN, and LSTM with the data numbers 1,000, 1,000, and 300, respectively. The training of these architectures is time-consuming. {In addition, the early-stopping strategy assumes that the performance ranking obtained by the partially-trained architectures is consistent with the actual performance ranking. However, the assumption is not always held because the convergence speed of different networks is not consistent as shown in Fig.~\ref{fig_learning_curve}. Accordingly, the early-stopping strategy may offer an inaccurate performance estimation. In fact, Zoph \textit{et al.}~\cite{zoph2018learning} also retrain the top-$250$ architectures with the highest performance until convergence on CIFAR-10 to discover the best architecture after searching. This confirmed the inaccuracy of the early-stopping strategy.}

\begin{figure}[!t]
	\centerline{\includegraphics[width=0.37\textwidth]{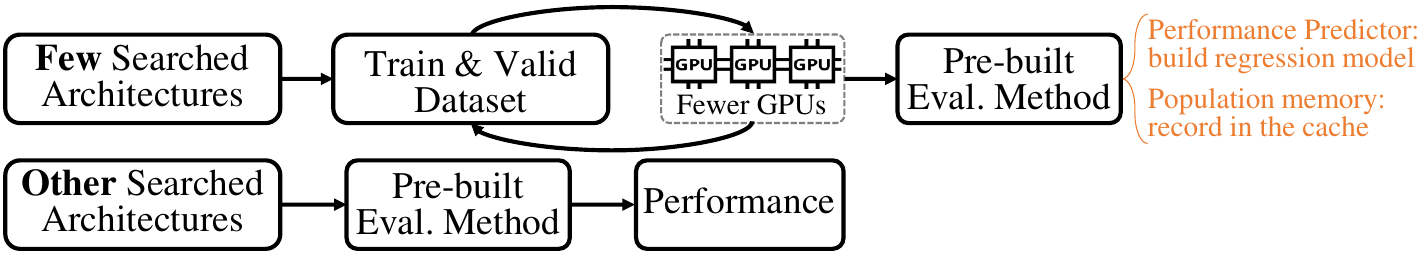}}
	{\caption{The flowchart of few-shot evaluation method and its relationship to subcategory methods. ``Eval.'' is the abbreviation of ``Evaluation''.}\label{few_shot_workflow}}
\end{figure}

\subsection{Few-shot Evaluation Methods}\label{few-shot}

As shown in Fig.~\ref{few_shot_workflow}, the few-shot evaluation methods first need to train a few architectures (i.e., $|\mathcal{A}^{EEM}|<|\mathcal{A}^s|$) to pre-build the evaluation method. Then, the performance of other architectures is obtained by the evaluation method. This can consume less runtime than TEM does.
According to the evaluation method utilized, existing few-shot methods can be divided into \emph{performance predictor} and \emph{population memory}. All these methods need to build the evaluation method and then use it to evaluate the performance of the searched architecture. Thus, their runtime contains initialization time and evaluation time.

\subsubsection{Performance Predictors}\label{pp}
The performance predictor is a kind of regression model, and can directly predict the performance of the architectures in $\mathcal{A}^s$ after being trained on the architecture set $\mathcal{A}^{EEM}={\{(A_i,P_i)\}}^{|\mathcal{A}^{EEM}|}_{i=1}$ where $A_i$ denotes the $i$-th architecture and $P_i$ represents its actual performance. Because $|\mathcal{A}^{EEM}|$ is generally less than $|\mathcal{A}^s|$ and the time of predicting the performance of the architectures in $\mathcal{A}^s$ is negligible, the performance predictors would consume less runtime than TEM. The main steps of building performance predictors are as follows: 
\begin{enumerate}[\textbf{Step} 1]
	\item Sample and train some architectures in the predefined search space to serve as the architecture set;\label{pp_step1}
	\item Encode the architectures in the architecture set;\label{pp_step2}
	\item Train a performance predictor to map the encoding of architectures to the corresponding performance values;\label{pp_step3}
\end{enumerate}

\textbf{Literature examples for each Step:}
We will separately discuss the existing works about steps~\ref{pp_step1}-\ref{pp_step3} in this part, and the keywords of the selected papers are ``performance predictor'' or ``neural predictor''.

\textbf{Step 1:}
The researchers often randomly sample a lot of architectures from the search space and train them from scratch to construct the architecture set. However, the set obtained by random sampling may not cover all representative architectures, which may result in the weak generalization of the performance predictor. Consequently, some works use other sampling methods to address this aspect. For example, Dai \textit{et al}.~\cite{dai2021fbnetv3} used Quasi Monte-Carlo to sample architectures from the search space. Hassantabar \textit{et al.}~\cite{hassantabar2022curious} chose Sobol sequences to sample architectures from the search space because it can generate more evenly distributed samples. The aims of these works are all to generate a small architecture set that can sufficiently represent the entire search space, thus minimizing the number of architectures required to be trained.

\textbf{Step 2:}
The architecture set gained in Step~\ref{pp_step1} cannot be directly fed into the regression model and must be encoded into a form that the regression model can tackle. In practice, there are two popular ways to encode the architectures: the sequence-based scheme and the graph-based scheme~\cite{ning2020generic}. 

The sequence-based scheme encodes the specific serialized information of the architectures and flattens the entire architectures to the strings. This kind of encoding scheme is broadly used in various search spaces. For example, Deng \textit{et al.}~\cite{deng2017peephole} employed a layer-based search space, proposed a uniform layer code to encode each layer to numerical vectors, and used the LSTM which is effective in processing the sequential data to integrate the information along a sequence of layers into a final string. Sun \textit{et al.}~\cite{sun2019completely} proposed a block-based search space that is composed of the ResNet block, DenseNet block, and pooling block, and encoded each block based on its kernel size, input channel, and output channel by sequence, and obtained a string at last. For the cell-based search space, Luo \textit{et al.}~\cite{luo2018neural} used the identifier of the input layers and the name of the operation applied to encode each layer to a string token. Then, a sequence of the discrete tokens was used to describe the given architectures. The sequence-based scheme is easy to implement and straightforward. However, it can only implicitly model the topological information of the given architectures. This makes performance predictors hard to capture topological information from the encoding. As a result, this scheme may lead to the reduction of the prediction ability of performance predictors for some architectures with rich topology information.

The graph-based scheme generally regards the architecture as the Directed Acyclic Graph (DAG). It explicitly represents the topological information of the architecture. For example, Wen \textit{et al.}~\cite{wen2020neural} used the adjacency matrix to describe the connections between operations and the one-hot codes to represent the operation type. To encode more information, Xu \textit{et al.}~\cite{xu2021renas} obtained the vectors of the operations, the FLating-point Operations Per second (FLOPs), and the parameter size for each operation. Then, these vectors were broadcasted into the adjacency matrix to generate the type matrix, the FLOP matrix, and the parameter matrix, respectively. At last, these matrices were concatenated as the final encoding. Ning \textit{et al.}~\cite{ning2020generic} modeled the operations as the transformation of the propagating information to mimic the data processing of the architecture. Although the graph-based scheme can better encode the topological information, it is not suitable for the layer-based and block-based search spaces which are almost the linear structure. The introduction of topological information may bring redundant information for them~\cite{luo2020accuracy}.

\textbf{Step 3:}
The performance predictor can be essentially treated as the regression model, and its training can be formulated as Equation~(\ref{eq_PP}):
\begin{equation}\label{eq_PP}
	\min _{T_{p}} \mathcal{L}\left(R\left(T_{p}, Encoder(X)\right), y\right)
\end{equation}
where $T_{p}$ is the trainable parameters of the regression model $R$, and $\mathcal{L}(\cdot)$ denotes the loss function. $Encoder(\cdot)$ denotes the encoding method. $X$ and $y$ correspond to the architecture and its performance in the architecture set, respectively. Then, we introduce the training process of the performance predictor from two aspects: the regression model and loss function.

In terms of the regression model, we illustrate it based on the encoding scheme because different models are good at handling different types of data. First, for the sequenced-based schemes, the models suitable for the sequential data are broadly used such as LSTM~\cite{deng2017peephole,istrate2019tapas}, RNN~\cite{liu2018progressive}, Transformer. Furthermore, some non-neural models are also used such as random forest and gradient boosting decision tree because of their ability to distinguish feature importance~\cite{luo2020accuracy}. As for the graph-based scheme, GNNs are commonly used because of its superiority in processing graph data~\cite{wen2020neural}. 

With regard to the loss function, it can be mainly divided into two types. The first is the element-wise loss function that calculates the distance between the predicted label and the ground-truth label for every sample. The most commonly used is Mean Square Error (MSE) shown as Equation~(\ref{eq_mse}):
\begin{equation}\label{eq_mse}
	\mathcal{L}_{MSE}=\frac{1}{N} \sum_{i=1}^{N}(y_{i}- f({x_{i}}))^{2}
\end{equation}
where $x_{i}$ and $y_{i}$ refer to the architecture and its corresponding performance, $N$ represents the number of samples. $f(\cdot)$ represents the performance predictor.
Huber loss is also a popular element-wise loss function as shown by Equation~(\ref{eq_huber}). It can prevent the model from being dominated by outliers~\cite{dai2021fbnetv3}.
\begin{eqnarray}\label{eq_huber}
  \mathcal{L}_{Huber} = \left\{\begin{array}{ll}
  \frac{1}{2}(y_{i}-f(x_{i}))^{2} & \text { If }|y_{i}-f(x_{i})| \leq \delta \\
  \delta|y_{i}-f(x_{i})|-\frac{1}{2} \delta^{2} & \text { otherwise }
  \end{array}\right.
\end{eqnarray}
In Equation~(\ref{eq_huber}), $\delta$ can be set according to the specific situation. When the prediction deviation is less than $\delta$, $\mathcal{L}_{Huber}$ uses the squared error. Otherwise, it uses linear error.
The second type of loss function is the pair-wise ranking loss function. It is firstly used in the performance predictor in~\cite{xu2021renas}. This is because the performance ranking is more important than the absolute performance of every architecture for NAS. The pair-wise ranking loss function just cares more about the performance ranking. Specifically, a typical hinge pair-wise ranking loss is represented by Equation~(\ref{eq_pw}):
\begin{eqnarray}\label{eq_pw}
  \mathcal{L}_{pairwise} = \frac{1}{N} \sum_{i = 1}^{N} \frac{1}{\left|T_{i}\right|} \sum_{j \in T_{i}} \operatorname{Max}\left(0, m-\left(y_{i}-y_{j}\right)\right)
\end{eqnarray}
where $T_{i} = \{j | y_{i} > y_{j}\}$. $m$ is the comparison margin.

\textbf{Literature examples based on the categorization:}
According to whether the performance predictors are retrained for updating during the process of NAS, the performance predictors can be divided into online and offline predictors.

For the offline predictor, the architecture-performance pairs have been completely collected before training the predictors. Note that the architecture set $\mathcal{A}^{EEM}$ of offline performance predictor is not sampled from the searched architecture $\mathcal{A}^s$. Once the predictors begin to work, there will be no new sample added to the training set. Many works employ this way to train the predictors owing to its simplicity. For example, Wen \textit{et al.}~\cite{wen2020neural} directly trained a small number of the architectures to train a GCN model as the performance predictor. However, the performance of offline predictors largely relies on the quality of the collected data. If the quality is poor, the predictor may also result in poor performance. Furthermore, once new samples are added to the dataset, the offline predictor has to be retrained, which makes the offline predictor inflexible. 

Different from offline predictors, online predictors can add new samples to improve their performance after they have been trained and used. As a result, it is more flexible and practical than offline predictors. Most online predictors first train a small number of architectures to build the predictor. Then, they choose some untrained architectures and predict their performance by predictors. Finally, the architectures with the top-$k$ highest predicted performance are trained and added to the architecture set to retrain the predictor. As a result, the architectures in $\mathcal{A}^{EEM}$ are generally partially sampled from $\mathcal{A}^s$ because the newly added architecture is usually found by the NAS algorithm. Please note the runtime of the online predictor also includes the update time, in addition to the initialization time and evaluation time mentioned in Subsection~\ref{org_pd}. Specifically, the update time refers to the time for retraining the offline predictor. Many works adopted online performance predictors. For instance, Liu \textit{et al.}~\cite{liu2018progressive} started the training of the predictor with the simple architectures, and continuously used the predictor to evaluate more complex architectures. Then, they picked and trained the most promising architectures as the new samples to finetune the predictor. Wei \textit{et al.}~\cite{wei2022npenas} proposed an EC-based search strategy, and trained the predictor with the architecture-performance pairs of the initial population. Then, they trained the architectures in the next population and added them into the architecture set to retrain the performance predictor.

Depending on if the architectures are fully labeled, the existing works can also be classified into supervised and semi-supervised performance predictors. 

The supervised performance predictors completely use the labeled architectures to train the predictors. Most predictors belong to the supervised one because it is simple. However, the number of labeled architectures is generally limited because the annotation of architectures is expensive. As a result, the supervised performance predictors generally try to extract more meaningful features from the limited architectures to improve their performance. For example, Chen \textit{et al.}~\cite{chen2021not} proposed an operation-adaptive attention module in the predictor to capture the relative significance between operations. Ning \textit{et al.}~\cite{ning2020generic} designed an encoding method that can model the calculation process of neural architectures to extract more meaningful information from architectures. 

Unlike the supervised predictors, the semi-supervised ones not only use the labeled architectures but also utilizes the unlabeled architectures. This is because the massive unlabeled architectures can provide invaluable information to optimize the performance predictor. For example, Tang \textit{et al.}~\cite{tang2020semi} used an AutoEncoder to learn meaningful features from both the labeled and unlabeled architectures and constructed a related graph to capture the inner similarities. Then, the related graph and the acquired features were mapped into the performance value by a GCN predictor. Luo \textit{et al.}~\cite{luo2020semi} trained the performance predictor with a small set of annotated architectures. Then, they used the predictor to predict the performance of unlabeled architectures. These predicted architectures were then used to retrain the performance predictors.

\textbf{Advantages:} Performance predictor has emerged as one of the research hotspots in EEMs in recent years~\cite{deng2017peephole,liu2018progressive,sun2019surrogate}. It has fast evaluation speed since the searched architecture does not need to be evaluated on GPUs.

\textbf{Disadvantages:} One critical problem for the performance predictor is that many fully-trained architectures are required to build the performance predictor. The contradiction of this problem lies in that training a large number of architectures may conflict with the design intention for EEMs, but a small number of architectures may lead to the overfitting problem~\cite{shorten2019survey}. In practice, the performance predictors always face a lack of data. The existing works solve the problem mainly from three aspects: encoding method, model, and data. In terms of the encoding method, researchers focus on representing more useful information about the architecture by the encoding method. For example, Xu \textit{et al.}~\cite{xu2021renas} proposed a graph-based encoding method to extract more valuable information about the architecture. Specifically, they encoded not only the topological information and operation type of each node but also the FLOPs and parameters of each node. As for the model, many works focus on how to extract more meaningful features from the encoding. For example, Wang \textit{et al.}~\cite{wang2019alphax} proposed a multi-stage MLP model that used different models to predict the architectures in different ranges of performance values. With regard to the data, Liu \textit{et al.}~\cite{liu2021homogeneous} proposed a homogeneous augmentation method for the architectures to generate a large amount of labeled data in an affordable way. Although many solutions are proposed, as a common problem for machine learning, the lack of labeled architectures cannot be solved from the root. As large language models continue to advance, the exploration of performance predictors can shift its focus toward harnessing self-supervised learning methods to create a large prediction model for neural architecture. This study will largely improve the generalization of the performance predictor and eliminate the necessity of constructing separate performance predictors for each search space through independent data collection.

\subsubsection{Population Memory}

\begin{figure}[!t]
  \centerline{\includegraphics[width=0.34\textwidth]{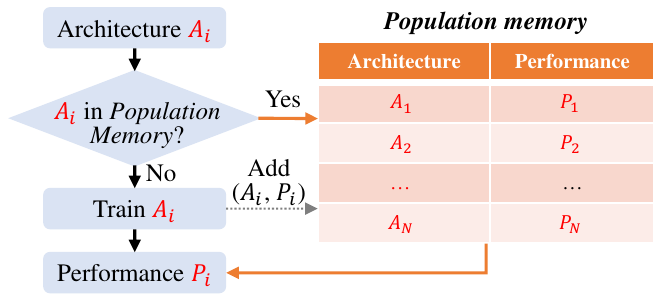}}
  \caption{A workflow of the population memory. Specifically, for each searched architecture $A_i$, if it is in the population memory that stores architectures and their corresponding performance, we can directly gain its performance $P_i$. Otherwise, it will be trained to obtain the performance $P_i$, and ($A_i$, $P_i$) will be subsequently stored in population memory.}\label{population_memory}
\end{figure}
Population memory is popular to accelerate fitness evaluation in EC. It is then used in the EC-based NAS to avoid evaluating the same architecture repeatedly. This is because the individuals in the previous populations may appear in the latter populations, and it is a waste of computational resources to reevaluate these individuals. The population memory can be seen as a cache system, and it mainly shortens time by reusing the architectural information that has appeared. For the population memory, the trained architecture set $\mathcal{A}^{EEM}$ is a subset of the searched architecture set $\mathcal{A}^s$ (i.e., $\mathcal{A}^{EEM}\in \mathcal{A}^s$). The workflow of the population memory is shown in Fig.~\ref{population_memory}. In this way, we can avoid evaluating the architectures repeatedly when facing the same architectures as the previous architectures. Many EC-based NAS adopted the method to avoid unnecessary costs. 

\textbf{Literature examples:} We utilize the ``population memory'' and  ``fitness memory'' as keywords to choose the specific papers.
Fujino \textit{et al.}~\cite{fujino2017deep} used a population memory to store the performance value of the individuals, and retrieved from memory when the architectures with the same encoding as the previous architectures appear. Similarly, Sun \textit{et al.}~\cite{sun2020automatically} designed a global cache system to record the hash code and the performance of the architectures. They obtained the performance when facing an architecture whose encoding has been stored in the cache.

\textbf{Advantages:} Compared with other EEMs, population memory allows for a more accurate evaluation of performance because the obtained performance by population memory is equal to the actual performance.

\textbf{Disadvantages:} The shortcoming of the population memory method is its lack of flexibility. Specifically, it can not be used to evaluate the architectures that have not been trained before. However, in practice, the repetitive architectures only account for a small portion of all architectures searched~\cite{sun2020automatically}.

\subsection{One-shot Evaluation Methods}\label{one-shot}

\begin{figure}[!b]	\centerline{\includegraphics[width=0.35\textwidth]{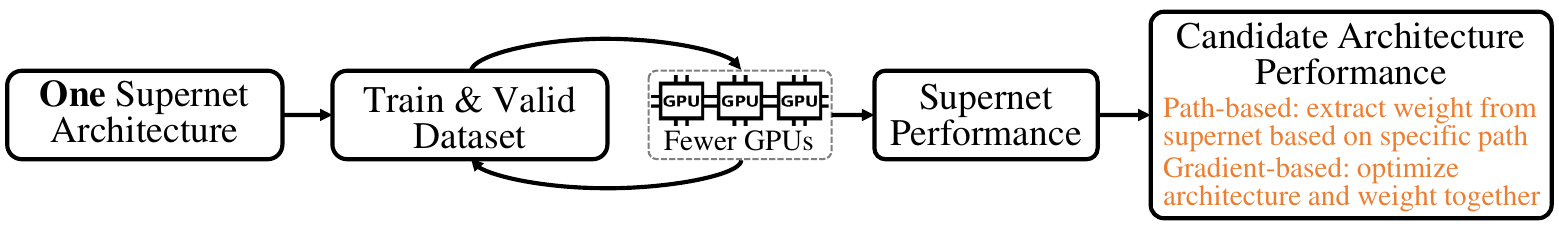}}
	{\caption{The flowchart of one-shot evaluation method and its relationship to subcategory methods.}\label{one_shot_workflow}}
\end{figure}

As indicated in Fig.~\ref{one_shot_workflow}, the one-shot method only requires training one neural architecture during the whole search process (i.e., $|\mathcal{A}^{EEM}|=1$). It is also called the weight-sharing method in the field of NAS. Because only one network (also called ``supernet'') needs to be trained, the method is cost-saving. Depending on whether architecture search and supernet training are coupled, the existing one-shot methods can be classified into \emph{path-based methods} and \emph{gradient-based methods}. All of them do not rely on the pre-build models, and the runtime only comprises evaluation time.

\subsubsection{Path-based Methods}\label{one-shot-path}
\begin{figure}[!b]
  \centerline{\includegraphics[width=0.40\textwidth]{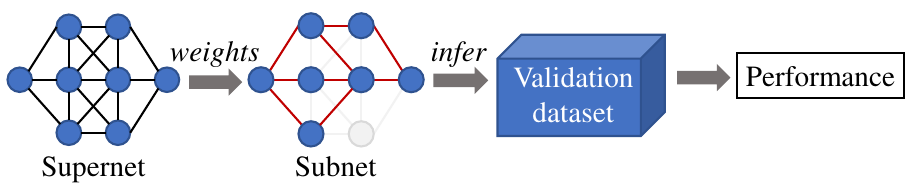}}
  \caption{The workflow of evaluating subnets for the path-based method. The grey lines and nodes in the subnet represent inactive connections and nodes, respectively. Only the weights of the supernet are obtained by training. The weights of the subnet are extracted from the supernet. Then, the performance of these subnets can be obtained by inferring on the validation dataset.}\label{fig_path-based}
\end{figure}

The path-based method trains a supernet that contains all candidate architectures (i.e., subnet). The weights of subnets can directly be extracted from the supernet. Each path refers to a subnet in the supernet. The workflow of evaluating subnets is shown in Fig.~\ref{fig_path-based}. The main workflow of the path-based method is as below:
\begin{enumerate}[\textbf{Step} 1]
	\item Design the supernet subsuming all the candidate architectures;
	\item Train the supernet from scratch by a path sampling strategy;\label{path-2}
	\item Predict the searched architectures by inheriting the weights from the supernet and inferencing the performance on the validation dataset.
\end{enumerate}

Among these steps, the training of the supernet (i.e., Step~\ref{path-2}) is critical. This is because the weights of the supernet are obtained through training and the weights of the subnet are extracted from the supernet. In practice, the weights of the supernet are almost not directly optimized. The reason is that the weights in the supernet are deeply coupled with the architecture of the supernet. Accordingly, the supernet is not robust to the changes in the architecture~\cite{bender2018understanding,guo2020single}. This may lead to a low correlation between the performance obtained by inheriting the weights and the actual performance.

\textbf{Literature examples:} Many path sampling strategies are proposed to decide which path(s) to train in every epoch of the supernet training to decouple the weights\&architecture of the supernet, and the keywords we used to select the studies are ``one-shot'' and ``single path''.  For example, Bender \textit{et al.}~\cite{bender2018understanding} proposed to randomly drop a subset of the operations in each iteration. The probability of the dropout is determined by a parameter called dropout rate. However, experiments in~\cite{bender2018understanding} indicated the training is sensitive to the dropout rate, which makes the training of supernet complicated. Guo \textit{et al.}~\cite{guo2020single} proposed to randomly sampled a subnet in the supernet, and only the weights of the subnet are updated in each iteration. This method is hyperparameter-free and thus avoids the defect of the path dropout strategy which is sensitive to the dropout rate.
Chu \textit{et al.}~\cite{chu2021fairnas} pointed out that the inherent unfairness in the method of Guo \textit{et al.} may lead to biased evaluation. They thus proposed strict fairness sampling strategy. This strategy samples $m$ ($m$ is the number of choice blocks per layer) at each epoch. In this strategy, all choice blocks are activated only once and are optimized only on one batch of data. 
Zhang \textit{et al.}~\cite{zhang2020one} designed the novelty-driven sampling strategy. Only the weights of the architectures sampled by novelty search were optimized. 
You \textit{et al.}~\cite{you2020greedynas} proposed a multi-path sampling strategy with rejection to filter the weak paths by evaluating them on a portion of the validation dataset. During the training of supernet, only those potential paths were trained. 
Zhang \textit{et al.}~\cite{zhang2023shiftnas} recognized that subnets with varying computational resources might require distinct training strategies. To address this, they proposed the probability shift, which is learned based on the training sufficiency of each subnet. This probability shift enables the automatic adjustment of sampling probabilities, optimizing the training process of the supernet.

\textbf{Advantages:} The path-based method only needs to train one architecture (i.e., $|\mathcal{A}^{EEM}|=1$) and the inference on the validation dataset is quick, thus it can largely accelerate the performance evaluation.

\begin{figure}[!t]
  \centerline{\includegraphics[width=0.33\textwidth]{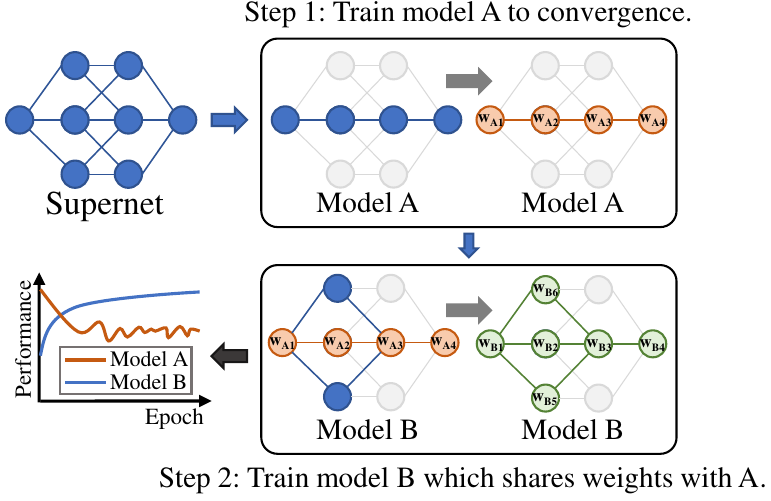}}
  \caption{An example of the multi-model forgetting problem. Model A and Model B are two different subnets that share part of the weights in the supernet. During the training of the supernet, model A is firstly trained in Step 1 and then the model B is trained in Step 2. This will cause the multi-model forgetting problem. Specifically, the weights of model A are optimized in the supernet at Step 1. Then, model B which shares partial weights with model A is trained at Step 2. During the training of model B, the performance of model A declines. The phenomenon is called multi-model forgetting.}\label{fig_multi_model}
\end{figure}

\textbf{Disadvantages:} Although the prediction accuracy of the path-based methods is improved with the help of the path sampling strategies, they still face the multi-model forgetting problem~\cite{benyahia2019overcoming}. Specifically, the multi-model forgetting problem refers to the performance decline of the previous-trained subnets when training subsequent subnets. This is because of the overwriting of shared weights when training the subsequent subnets with partially-shared weights one by one. An example of this problem is shown in Fig.~\ref{fig_multi_model}. The multi-model forgetting problem will deteriorate the prediction ability. As a result, although the path-based methods can greatly reduce computational time, they may result in unreliable performance ranking. Many solutions are proposed to alleviate the multi-model forgetting problem. For example, Benyahia \textit{et al.}~\cite{benyahia2019overcoming} proposed a statistically-justified weight plasticity loss to regularize the optimization of the shared parameters. In this way, they can preserve the shared parameters that were important for the previous model. Zhang \textit{et al.}~\cite{zhang2020forgetting} formulated the supernet training as a constrained continual learning optimization problem. During the training of the current subnet, they constrained the training loss of all previous subnets is less than that in the last step. However, the number of constraints increases linearly as the increase of previous subnets. Thus, they proposed a greedy novelty search method to select a subset of existing constraints from previous subsets to overcome the multi-model forgetting problem.
Although the above works can alleviate the phenomenon of multi-model forgetting, they cannot solve the problem from the root. This is because the problem is caused by the mechanism the path-based method trains the supernet to evaluate subnets. As a result, the ranking correlation between the performance obtained by the path-based method and the actual performance may be poor.

\subsubsection{Gradient-based Methods}\label{one-shot-gradient}
Unlike the path-based methods, the gradient-based methods couple the supernet training and architecture searching. Specifically, they relax the discrete search space to be continuous, and jointly optimize the architecture and the supernet weights by the gradient-based search strategy. The most famous work falling into this category is DARTS~\cite{liu2018darts} proposed in 2019. Then, the DARTS approach became the dominant approach to gradient-based methods owing to its effectiveness. Specifically, DARTS used a cell-based search space and regarded the supernet as a DAG with $N$ nodes. Each node $x^{(i)}$ is a latent representation such as the feature map, and each edge $(i,j)$ refers to the operation $o^{(i,j)}$ that can transform $x^{i}$. Each intermediate node is computed based on all of its predecessors as shown in Equation~(\ref{eq_pre}):
\begin{eqnarray}\label{eq_pre}
  x^{(j)} = \sum_{i<j} o^{(i, j)}\left(x^{(i)}\right)
\end{eqnarray}
To relax the discrete search space to be continuous, the categorical choice of a particular operation is continuously relaxed by employing a softmax function over all the possible operations as shown in Equation~(\ref{eq_op}):
\begin{eqnarray}\label{eq_op}
  \bar{o}^{(i, j)}(x) = \sum_{o \in \mathcal{O}} \frac{\exp \left(\alpha_{o}^{(i, j)}\right)}{\sum_{o^{\prime} \in \mathcal{O}} \exp \left(\alpha_{o^{\prime}}^{(i, j)}\right)} o(x)
\end{eqnarray}
where $\mathcal{O}$ is a set of candidate operations. $\alpha^{(i,j)}$ denotes the operation mixing weights (i.e., architecture parameter) for a pair of nodes $(i,j)$. An architecture can be obtained by replacing each mixed operation $\overline{o}^(i,j)$ with the most likely operation, i.e., $o^{(i, j)}=\operatorname{argmax}_{o \in \mathcal{O}} \alpha_{o}^{(i, j)}$. Then, DARTS aims to jointly learn the architecture parameters $\alpha$ and the weights $\omega$ which determine the training loss $\mathcal{L}_{train}$ and validation loss $\mathcal{L}_{val}$. This can be formulated as a bilevel optimization with $\alpha$ as the upper-level variable while $\omega$ as the lower-level variable problem, modeled by Equation~(\ref{eq_biop}):
\begin{eqnarray}\label{eq_biop}
\left\{\begin{matrix}
  \min _{\alpha} \mathcal{L}_{val}\left(w^{*}(\alpha), \alpha\right) \\
  \text { s.t. } w^{*}(\alpha)=\operatorname{argmin}_{w} \mathcal{L}_{train}(w, \alpha)
  \end{matrix}\right.
\end{eqnarray}
By this way, we can find $\alpha^*$ that minimize the $\mathcal{L}_{val}(w^{*}(\alpha),\alpha)$ where the weights $\omega^{*}$ minimize the $\mathcal{L}_{train}(w, \alpha)$.

\textbf{Literature examples:} 
{Many gradient-based methods are proposed to improve effectiveness, and we utilize ``differentiable NAS,'' ``differentiable architecture search,'' and ``differentiable neural architecture search'' as the keywords to choose the papers. These papers mainly focus on the three research directions as follows:}

\begin{itemize}
    \item {\textbf{Reduce memory:}} {Many works try reduce the memory of gradient-based methods.} For example, Cai \textit{et al.}~\cite{cai2018proxylessnas} leveraged binarized architecture parameters to guarantee only one path of activation is active in memory at runtime. Xu \textit{et al.}~\cite{xu2019pc} randomly sampled a subset of channels as the proxy of all the channels to reduce memory consumption. In addition, they used edge normalization, which adds a new set of edge-level parameters to improve the stability of the search. Similarly, Xue \textit{et al.}~\cite{xue2022partial} saved the memory by adopting the subset of the channels. The difference is that they proposed an attention mechanism, and selected the channels with higher attention weights. The method can better transmit important feature information into the search space and prevent the instability of the search.
    \item {\textbf{Enhance generalization:}} The gradient-based method generally suffers from poor generalization on various datasets of the searched architectures. To solve this, Li \textit{et al.}~\cite{li2020adapting} introduced the idea of domain adaptation. They improved the generalizability of architectures by minimizing the generalization gap between domains. Liu \textit{et al.}~\cite{liu2021mixsearch} proposed a novel method to mix the data from multiple tasks and domains into a composited dataset. Ye \textit{et al.}~\cite{ye2022beta} proposed a regularization method that can maintain a small Lipschitz of the searched architectures. This is because the architectures have good generalization ability if its Lipschitz is small.
    \item {\textbf{Improve stability:}} {The gradient-based method lacks stability because of performance collapse. Existing works to alleviate this problem can be divided into five categories. The first category typically uses dropout or stop operations to limit the number of skip connections, such as P-Darts~\cite{chen2019progressive} and DARTS+~\cite{liang2019darts+}. However, they would reject some potentially high-performance architectures with multiple skip connections. The second category refers to regularizing relevant indicators, such as Hessian eigenvalues~\cite{arber2020understanding,chen2020stabilizing} and layer alignment~\cite{movahedi2022lambda}, to indicate the performance collapse of the network. The main limitation of these methods is their reliance on the quality of the indicator. The third category is known as modifying the cell, which often involves auxiliary skip connection~\cite{chu2020darts} and additional sigmoid function~\cite{chu2020fair}. Thus, they are computationally expensive. The fourth category is regarded as dropping out unimportant operations based on the group during the training process~\cite{hong2022dropnas,gu2021dots}. Unfortunately, they highly rely on the manually set hyperparameters, e.g., group number. The fifth catagory is based on the regularization method, such as Beta-decay~\cite{ye2022beta} and BatchNorm of the supernet~\cite{zhang2023differentiable}. However, they may lead to overly conservative updates, preventing the ability to explore the search space.}
\end{itemize}

\textbf{Advantages:} The gradient-based method combines supernet training and architecture searching, which is more effective than the path-based method and achieves great performance on a large-scale dataset.

\textbf{Disadvantages:} There are three disadvantages of the gradient-based method: (1) Huge memory: The gradient-based methods always suffer from large memory because of the joint optimization which not only trains a supernet but also searches for the architectures. (2) Low generalization: The architecture search by radient-based method cannot perform as well as the dataset used in the search on other datasets. (3) Performance collapse: The gradient-based methods, e.g., DARTS, tend to accumulate parameter-free operations (i.e., skip connection) which lead to rapid gradient descent~\cite{arber2020understanding,chu2020darts}. However, learnable operations such as convolution are the better choice for improving the expression ability of DNNs. As mentioned above, existing works can only alleviate these problems but cannot address them.

In summary, the one-shot method can largely reduce the computational time by only training one supernet. However, the path-based methods experience the phenomenon of multi-path forgetting, and the gradient-based method faces the problem of performance collapse. These problems may lead to inaccurate ranking. Hence, the one-shot method may fail to reflect the true ranking of architectures. This also affects the performance of the discovered architecture. The evidence is that the architecture searched by the one-shot method performs sometimes worse than those discovered by random search~\cite{yang2019evaluation}.

\subsection{Zero-shot Evaluation Methods}\label{zero-shot}
\begin{figure}[!b]	\centerline{\includegraphics[width=0.33\textwidth]{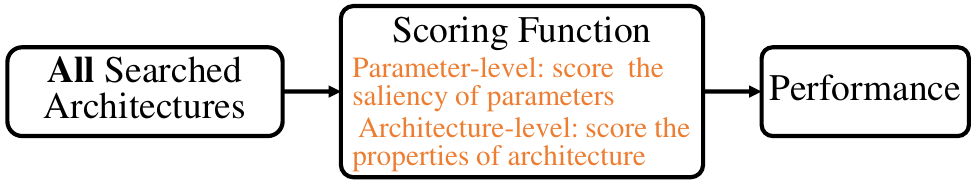}}
	{\caption{The flowchart of zero-shot evaluation method and its relationship to subcategory methods.}\label{zero_shot_workflow}}
\end{figure}

For all searched architectures, they do not need to be trained on GPUs (i.e., $\mathcal{A}^{EEM}=\emptyset$), instead of evaluating them by scoring function as shown in Fig.~\ref{zero_shot_workflow}. Based on the type of scoring function, the zero-shot methods can be mainly classified into \emph{parameter-level methods} and \emph{architecture-level methods}. In addition, the scoring function is typically a mathematical formula that does not rely on pre-built evaluation methods. Thus, the runtime of the zero-shot method is the evaluation time to obtain the score. Because getting the score of architecture is often time-saving, it has an extremely low cost. Please note that zero-shot methods cannot directly evaluate the absolute performance of a given architecture. It can only rank the architectures based on the score of the performance. 

\subsubsection{Parameter-level Methods}\label{zero_shot_parameter}
The parameter-level methods first use the indicator to measure the saliency (i.e., importance) of certain parameters. Then, they score for the entire architecture by aggregating the saliencies of certain parameters. The methods only required a minibatch of data and a forward/backward propagation pass to calculate the indicators for certain parameters. Because it involves no training, it can consume much less runtime than TEM. Generally, the saliency indicators generally come from the network pruning literature. They are originally used to evaluate the importance of parameters and thus remove the unimportant ones. 

\begin{table*}[ht]
\caption{Summary of different EEMs in NAS including their principle and relative references. The first column represents the categories of EEMs. {The second column illustrates the subcategories of each category with their runtime composition, where `IT' and `ET' denote the initialization time and evaluation time, respectively. The fourth and fifth columns provide the advantages and disadvantages of each method, respectively.} The sixth column describes the working principles associated with each subcategory. The fifth column indicates the relevant references corresponding to each subcategory. }
\begin{center}
\begin{tabularx}{\textwidth}{p{0.8cm}p{2.2cm}p{4cm}p{2.8cm}Xp{2.5cm}}
\Xhline{1.5pt}
Category & Subcategory & \makecell[c]{Working Principle} & {\makecell[c]{Advantages}} & {\makecell[c]{Disadvantages}} & \makecell[c]{References} \\ 
\Xhline{1pt}
\multirow{8}{*}{$N$-shot}  & \multirow{2}{*}{\shortstack{Downscaled dataset \\{(ET)}}}   & Training time reduced by training on the down-sampled dataset.  & {\multirow{2}{*}{Easy to implement}} & {Ranking mismatch on the downscaled datasets} & \cite{zoph2018learning,liu2019deep,chrabaszcz2017downsampled,hassantabar2022curious,klein2017fast,shu2020automatically,sapra2020constrained,wang2020particle,xu2021partially,liu2017structure} \\  
 & \multirow{2}{*}{\shortstack{Downscaled model \\{(ET)}}} & Training time reduced by downscaling architectures. & {\multirow{2}{*}{Save GPU memory}} &  {Low correlation between original and proxy models} & \cite{zoph2018learning, real2019regularized, liu2018darts} \\ 
  & \multirow{2}{*}{\shortstack{Network morphism \\ {(ET)}}} & Training time reduced by inheriting weights from the parent model. & {Leverage existing knowledge} & {Failed when child network not exist in parent network} & \cite{cai2018efficient,elsken2018efficient,chen2015net2net,wei2016network,cai2018path,elsken2017simple,zhang2022self}\\
 & \multirow{2}{*}{\shortstack{Learning curve \\ extrapolation {(IT, ET)}}} & The performance is extrapolated after only training for a few epochs. & {Easily combine with various NAS algorithms}  & {Need trained network to build the prediction models} & \cite{zoph2018learning,baker2017accelerating,klein2016learning,domhan2015speeding,rawal2018nodes,kim2022two}  \\  
 \hline
\multirow{5}{*}{Few-shot}  & \multirow{2}{*}{\shortstack{Performance predictor \\ {(IT, ET)}}}  & Use a regression model to directly predict the performance of DNNs & {Fast GPU-freed evaluation speed} & {Need fully-trained architectures to build the regression model} & \cite{rawal2018nodes,baker2017accelerating,deng2017peephole,liu2018progressive,sun2019surrogate,istrate2019tapas,tang2020semi,luo2020accuracy,sun2021novel,huang2022arch,white2021bananas,dudziak2020brp,chen2021contrastive,hassantabar2022curious,mauch2020efficient,dai2021fbnetv3,liu2021fox,li2021generic,li2020gp,wen2020neural,lu2021tnasp,chen2021not,wei2022npenas,peng2022pre,wang2021rank,xu2021renas,wu2021stronger} \\ 
 & \multirow{2}{*}{\shortstack{Population memory \\ {(IT, ET)}}} & Store the trained DNNs and directly match them when used. & {Loss-less evaluation performance} & {Lack of flexibility cannot evaluate unseen architecture} & \cite{xie2022benchenas,sun2020automatically,fujino2017deep,johner2019efficient,miahi2022genetic}                   \\ \hline
\multirow{6}{*}{One-shot} & \multirow{3}{*}{\shortstack{Path-based \\ {(ET)}}} & Only the supernet requires being trained, where the weights of candidate architectures are from it. & {Fast, inference to get the performance of searched network} & {Multi-model forgetting problem deteriorates the prediction ability} & \cite{bender2018understanding,chen2019detnas,guo2020single,peng2020cream,huang2021searching,xia2022progressive,guo2020powering,liang2021opanas,chen2021one,dong2019one,zhang2020one,cai2020once,yan2021lighttrack,chen2021iterative,li2020improving,you2020greedynas,chu2021fairnas,shi2020bridging,wang2021attentivenas,liu2020comprehensive,zhang2023shiftnas}\\  
 & \multirow{3}{*}{\shortstack{Gradient-based \\ {(ET)}}} & Jointly optimize the architecture parameter and the supernet weights by bilevel optimization. & {Effectively, suitable for large-scale datasets} & {(1)Huge memory; (2) Low generalization; (3) Performance collapse} & \cite{zhang2023differentiable,ye2022beta, liu2018darts,wang2022eautodet,xu2022dnas,yu2022cyclic,zhang2022balenas,huang2022greedynasv2,dong2019searching,zhang2020overcoming,zhang2020forgetting,pan2022distribution,arber2020understanding,akimoto2019adaptive,fil2021darts,chu2020darts,liang2019darts+,gu2021dots,hong2022dropnas,zhou2021ec,chu2020fair,wu2019fbnet,wan2020fbnetv2,nayman2021hardcore,zhang2021idarts,xue2021idarts,xu2019pc,tang2021probeable,chen2019progressive,cai2018proxylessnas,xue2021rethinking,hou2021single,chen2020stabilizing,li2021one,movahedi2022lambda} \\ \hline
\multirow{6}{*}{Zero-shot} & \multirow{3}{*}{\shortstack{Parameter-level \\ {(ET)}}} & Aggregate the saliencies of all parameters to measure the entire DNN without training. & {Low computational complexity} & {\multirow{3}{*}{Unreliability and inaccurate}} & \cite{abdelfattah2020zero,shu2022unifying,li2022zico} \\  
 & \multirow{3}{*}{\shortstack{Architecture-level \\ {(ET)}}} & Measure the properties positively related to architecture performance without training. & {Consider the global characteristics of the architecture} & {\multirow{3}{*}{Inaccurate performance ranking}} & \cite{mellor2020neural,mellor2021neural,lin2021zen,chen2021neural,shu2021nasi,xu2021knas,zhou2022training}  \\ 
\Xhline{1.5pt}
\end{tabularx}
\end{center}\label{categories_AC}
\end{table*}

\textbf{Literature examples:} There have been some works proposed in this category, and the keywords of them are ``zero-shot'', and ``parameter saliency''. For example, Abdelfattah \textit{et al.}~\cite{abdelfattah2020zero} adopted a series of pruning-at-initialization metrics includes grad\_norm, snip~\cite{lee2018snip}, grasp~\cite{wang2020picking}, fisher~\cite{theis2018faster,turner2019blockswap}, synflow~\cite{tanaka2020pruning} to measure the saliency of parameters. Then, they summed up the metric value of all parameters to score for the architecture performance. 
Specifically, grad\_norm uses the Euclidean norm of the gradient after a forward/backward propagation pass of a minibatch of training data to measure the parameters. Snip calculates the saliency of a specific parameter by approximating the change in cross-entropy loss when the parameter is pruned. Synflow is a modified version of snip with the change that the loss in synflow is the product of all parameters in the network. Grasp calculates the saliency by measuring the approximate change in gradient norm when the parameter is removed. Fisher measures the approximate loss change when activation channels are removed. Li \textit{et al.}~\cite{li2022zico} discovered that a high-performance network tends to possess high absolute mean values and low standard deviation values for the gradient. Thus, they introduced the zero-shot inverse coefficient of variation, which considers both the absolute mean and standard deviation values of each layer.

\textbf{Advantages:} The parameter-level method has a low computational complexity because only a few simple parameters need to be computed.

\textbf{Disadvantages:} Despite the cost-saving of parameter-level methods, they are proven inaccurate. Simply summing up the saliencies of each parameter to measure the performance of architectures is problematic. This is because the saliency only reflects the impact of the parameter on the architecture. In fact, a recent work~\cite{ning2021evaluating} also pointed out that the existing parameter-level methods are not suitable for ranking architectures. Their ranking qualities even cannot surpass those of parameter size or FLOPs. This proves the unreliability of the parameter-level methods on the other hand.

\subsubsection{Architecture-level Methods}\label{zero_shot_architecture}
The architecture-level methods score the architectures by measuring the properties positively related to architecture performance. Because they do not need to train any architecture, they can largely save time for performance evaluation. 

\textbf{Literature examples:} Many architecture-level methods have been proposed to efficiently evaluate the performance of the architecture, and we use the keywords ``zero-shot,'' ``scoring at initialization,'' and ``pruning at initialization'' to select them. For example, Mellor \textit{et al.}~\cite{mellor2020neural, mellor2021neural} proposed to quantify the activation overlap of different inputs for a network to score the untrained networks. The learnability of architectures is more strong when the activation for different inputs is well separated. They proposed an indicator based on input Jacobian correlation in~\cite{mellor2020neural}. Furthermore, they also devised using the Hamming distance to judge how dissimilar the two inputs are in another version of this paper~\cite{mellor2021neural}. Xu \textit{et al.}~\cite{xu2021knas} assumed that the gradients can be used to evaluate the random-initialized networks. This is based on the observation that gradients can directly decide the convergence and generalization results. 
They presented the gradient kernel to take each layer as a basic unit. After that, the mean of the Gram matrix for each layer was computed to score the network. 
Lin \textit{et al.}~\cite{lin2021zen} averaged the Gaussian complexity of linear function in each linear region to measure the network expressivity.
Zhou \textit{et al.}~\cite{zhou2022training} observed the synaptic diversity of multi-head self-attention in Vision Transformer (ViT) affects the performance notably. Thus, they proposed an indicator to rank ViT architectures from the perspectives of synaptic diversity and synaptic saliency. Chen \textit{et al.}~\cite{chen2021neural} measured the architecture performance by analyzing the spectrum of the neural tangent kernel and the number of linear regions in the input space. 
Shu \textit{et al.}~\cite{shu2021nasi} proposed an efficient approximation of the trace norm of NTK to alleviate the prohibitive cost of computing NTK and estimate the performance of neural architectures.

\textbf{Advantages:} The architecture-level method can comprehensively evaluate the global characteristics of the architecture, such as hierarchy and connectivity. This helps capture the impact of architecture on performance.

\textbf{Disadvantages:} The architecture-level methods are usually motivated by some theoretic studies on neural networks. They proposed indicators to judge the trainability, learnability, generalization, or expressivity of the architectures and ranked the architectures. Despite their efficiency, the architecture-level methods always lead to inaccurate performance ranking. This is because they only roughly measure the properties of architectures that are positively related to architecture performance.

In summary, the zero-shot methods can further reduce the computational time compared with other methods. However, the performance of the zero-shot method is usually not good enough in practice. Moreover, the robustness of the zero-shot methods can not be guaranteed. The performance fluctuated dramatically among different tasks. 

So far, all categorizations of EEMs have been described in detail, and we have summarized these EEMs in Table.~\ref{categories_AC}. It is worth noting that there is a method with more than one EEM, also known as the hybrid method. For example, Zoph \textit{et al.}~\cite{zoph2018learning} combined a downsampled dataset and a downsampled model to decrease the evaluation time. Based on this, Zoph \textit{et al.}~\cite{zoph2018learning} even added the learning curve extrapolation methods to further improve the evaluation efficiency. In addition, Rawal \textit{et al.}~\cite{rawal2018nodes} introduced a performance predictor with the early stopping strategy to accelerate the evaluation process. Moreover, EEMs also have the potential to be combined with each other, e.g., performance predictors and population memory, and users could use them in practice as they need.

\begin{table*}[!t]
  \caption{The characteristics of NAS benchmarks. $|V|$ refers to the number of nodes, and $|E|$ denotes the number of edges. Please note that NAS-Bench-301 provides 60,000 annotated architectures and a performance predictor which can predict all $10^{18}$ architecture in the DARTS search space.}
  \begin{center}
\begin{tabularx}{\textwidth}{ccccccc}
\Xhline{1.5pt}
Benchmark     & Year & Task                           & Dataset                              & Search space  & OON/OOE &\#Architecture                                                                                                                                    \\ 
\Xhline{0.75pt}
NAS-Bench-101 & 2019         & Image classification           & CIFAR-10                             & $|V|\leq 7$,\#ops=3,$|E|\leq 9$&OON& 423,624  \\ 
NAS-Bench-201 & 2020         & Image classification           & \makecell[c]{CIFAR-10, CIFAR-100,\\ ImageNet-16-120} & $|V|=4$,\#ops=5 &OOE& 15,625    \\ 
NAS-Bench-301 & 2020         & Image classification           & CIFAR-10                             & $|V|\leq 7$,\#ops=7         &OOE& 60,000/$10^{18}$ \\ 
NAS-Bench-NLP & 2020         & Natural language processing    & Penn Tree Bank                       & $|V|\leq 24$,\#ops=7 &OOE& 14,322       \\ 
NAS-Bench-ASR & 2021         & Automatic Speech Recognization & TIMIT audio dataset                  & $|V|\leq 4$,\#ops=7,$|E|\leq 9$ &OOE& 8,242       \\ 
\Xhline{1.5pt}
\end{tabularx}
  \end{center}\label{tbl_benchmark}
  \end{table*}

\section{Evaluation}\label{evaluation}
In this section, we first discuss the evaluation metrics in Section~\ref{eva_metric}. Then, we introduce the commonly-used benchmark datasets in Section~\ref{eva_benchmark}. Finally, we report the performance of various EEMs on the benchmark datasets in Section~\ref{eva_comp}.

\subsection{Evaluation Metrics}\label{eva_metric}
Based on the no-free-lunch theory, the EEMs may perform inaccurately compared with TEM though their speed is much faster. As a result, we mainly measure the EEMs in two folds: runtime and prediction ability.
The runtime of an EEM is generally measured by GPU day (GPU day = The number of GPUs $\times$ The number of days)~\cite{sun2019completely}.
As for the measurement of prediction ability, the commonly used metrics can be mainly divided into correlation-based, ranking-based, and top-based. For the convenience of discussion, we denote the ground-truth performance and the predicted performance of architectures ${\{A_{i}\}}_{i=1}^{N}$ as ${\{y_{i}\}}_{i}^{N}$ and ${\{\hat{y}_{i}\}}_{i}^{N}$, respectively. The real performance ranking of the architecture $A_{i}$ (i.e., the ranking of $y_{i}$ among ${\{y_{i}\}}_{i}^{N}$) is $r_{i}$, while the predicted performance ranking for $A_{i}$ (i.e., the ranking of $\hat{y}_{i}$ among ${\{\hat{y}_{i}\}}_{i}^{N}$) is $\hat{r}_{i}$.

The correlation-based metric mainly calculates the correlation between the ground-truth performance and the predicted performance. It mainly includes the Pearson coefficient and the Coefficient of Determination (i.e., $\mathcal{R}^{2}$).
The Pearson coefficient is formulated by Equation~(\ref{eq_pearson}):
\begin{eqnarray}\label{eq_pearson}
  \mathcal{M}_{pearson} = \frac{\sum_{i = 1}^{N}\left(y_{i}-\bar{y}\right)\left(\hat{y}_{i}-\bar{\hat{y}}\right)}{\sqrt{\sum_{i = 1}^{N}\left(y_{i}-\bar{y}\right)^{2} \sum_{i = 1}^{N}\left(\hat{y}_{i}-\bar{\hat{y}}\right)^{2}}}
\end{eqnarray}
where $y_i$ and $\hat{y}_i$ represent the ground-truth and predicted performance, respectively. $\bar{y}$ and $\bar{\hat{y}}$ represent the average of all ground-truth performance and predicted performance, respectively. The $\mathcal{R}^{2}$ is formulated by Equation~(\ref{eq_r2}):
\begin{equation}\label{eq_r2}
\mathcal{M}_{R^{2}}=1-\frac{\sum_{i=1}^{N}\left(y_{i}-\hat{y}_{i}\right)^{2}}{\sum_{i=1}^{N}\left(y_{i}-\bar{y}_{i}\right)^{2}}
\end{equation}
For both metrics, the closer to one the value is, the more accurate the EEM is. 

The ranking-based metric calculates the correlation between the real performance ranking and the predicted one. 
Kendall's Tau ranking correlation (KTau)~\cite{sen1968estimates} and Spearman's ranking correlation (SpearmanR)~\cite{hauke2011comparison} are frequently-used ranking-based metrics. The KTau is the relative difference of concordant and discordant pairs, and is shown in Equation~(\ref{eq_Ktau}):
\begin{equation}\label{eq_Ktau}
\mathcal{M}_{KTau}=2 \times \frac{\text { number of concordant pairs }}{N(N-1) / 2}-1
\end{equation}
where the concordant pair means that the predicted rankings and the ground-truth rankings of a given pair are the same.
On the other hand, the SpearmanR is calculated based on the difference between the predicted ranking and the real ranking. It is shown in Equation~(\ref{eq_SpearmanR}):
\begin{eqnarray}\label{eq_SpearmanR}
  \mathcal{M}_{SpearmanR} = 1-\frac{6 \sum_{i = 1}^{N} (r_i-\hat{r}_i )^{2}}{N\left(N^{2}-1\right)}
\end{eqnarray}
For both KTau and SpearmanR, the value closer to one means that the EEM is more accurate. 

The top-based metric measures the ability of EEMs to discover the best architecture, which is important for NAS. The top-based metric is mainly composed of N@K and Precision@K. In specific, N@K refers to the best ground-truth ranking among the predicted top-K architectures. The value of N@K is a positive integer, and greater than or equal to one. A lower value of N@K means that the EEM is more capable of finding the best architecture. The Precision@K is the proportion of the ground-truth top-K architectures in the predicted top-K proportion architectures. Different from N@K, the value closer to one means the EEM is better at discovering best architectures.

\begin{table*}[!t]
  \caption{The Kendall's Tau (KTau) value of $N$-shot, few-shot, one-shot and zero-shot methods on NAS-Bench-101 (NB101), NAS-Bench-201 (NB201), and NAS-Bench-301 (NB301). Specifically, the first column shows the categories of EEMs, and the second column shows the subcategories of each category. The third column shows the specific method, and the fifth, sixth and seventh columns show the KTau of each specific method on NB101, NB201 and NB301 respectively. The eighth column is the experimental notes for obtaining KTau value. $^\dag$ denotes that the benchmark does not provide sufficient data to produce the experiments. $^\ddag$ indicates the core code of the method is not open-source and the results is absence in the original paper. $^{*}$ indicates NB301 does not provide the real accuracy value, so the community of performance performance would not provide the KTau value.}
  \begin{center}
    \begin{tabularx}{\textwidth}{c|c|ccccX}
      \Xhline{1.5pt}
      Category &  Subcategory & Method & NB101 & NB201 & NB301 & \makecell[c]{Note}\\ \Xhline{0.75pt}
      \multirow{8}{*}{$N$-shot} & \multirow{4}{*}{Downscaled dataset} & Proxy dataset & {-$^{\dag}$} & 0.827 & {-$^{\dag}$} & CIFAR-10 is proxy dataset, and ImageNet is original dataset. {$^{\dag}$ NB101 and NB201 only have CIFAR10 results.} \\ \cline{3-7} 
     & & Subset of dataset & {-$^{\dag}$} & 0.867 & {-$^{\dag}$} & Use CIFAR-10 as the subset of CIFAR-100. {$^{\dag}$ NB101 and NB201 only provide CIFAR10 results.} \\ \cline{2-7} 
     & \multirow{4}{*}{\shortstack{Learning curve \\ extrapolation}} & Early stopping(1/4) & {-$^{\dag}$}  & 0.602  & 0.614 &  Terminate training when the epoch is one quarter of the fully trained epoch. {$^{\dag}$ NB101 lacks 1/4 epoch results.} \\ \cline{3-7} 
     &   & Early stopping(1/2) & 0.438 & 0.663 & 0.662 & Terminate training when the epoch is one half of the fully trained epoch. \\ \hline
      \multirow{9}{*}{Few-shot} & \multirow{9}{*}{Performance predictor} & NeuralPredictor~\cite{wen2020neural} & 0.679 & 0.646 & {-$^{*}$} & 424 and 1,564 samples are used as the training data in NAS-Bench-101 and NAS-Bench-201, respectively. \\ \cline{3-7} 
     & & Peephole~\cite{deng2017peephole} & 0.4556 & {-$^{\ddag}$} & {-$^{*}$} & 424 samples are used as the training data.\\ \cline{3-7} 
      & & E2EPP~\cite{sun2019surrogate} & 0.5038 & {0.7669} & {-$^{*}$} & 424 samples are used as the training data. \\ \cline{3-7} 
     & & ReNAS~\cite{xu2021renas} & 0.657 & {-$^{\ddag}$} & {-$^{*}$} & 424 samples are used as the training data. \\ \cline{3-7} 
     & & HOP~\cite{chen2021not} & 0.813 & 0.897 & {-$^{*}$} & 381 and 781 samples are used as the training data in NAS-Bench-101 and NAS-Bench-201, respectively. \\ \cline{3-7} 
     & & TNASP~\cite{lu2021tnasp} & 0.722 & 0.726 & {-$^{*}$} & 424 and 1,564 samples are used as the training data in NAS-Bench-101 and NAS-Bench-201, respectively. \\ \cline{1-7} 
      \multirow{2}{*}{One-shot} & \multirow{2}{*}{Path-based} & OSNAS~\cite{ning2021evaluating} & 0.446 & 0.744 & 0.548 & Using MC sample in supernet training. \\ \cline{3-7} 
     & & FairNAS~\cite{chu2021fairnas} & {-$^{\ddag}$} & 0.706 & 0.527 & - \\ \hline
      \multirow{8}{*}{Zero-shot} & \multirow{6}{*}{Parameter-level} & synflow~\cite{abdelfattah2020zero} & -0.063 & 0.573 & 0.201 & - \\ \cline{3-7} 
     & & grad\_norm~\cite{abdelfattah2020zero} & -0.276 & 0.401 & 0.070 &  - \\ \cline{3-7} 
     & & snip~\cite{abdelfattah2020zero} & -0.206 & 0.402 & 0.050 & - \\ \cline{3-7} 
     & & grasp~\cite{abdelfattah2020zero} & -0.266 & 0.348 & 0.365 & - \\ \cline{3-7} 
     & & fisher~\cite{abdelfattah2020zero} & -0.202 & 0.362 & -0.158 & - \\ \cline{3-7} 
     & & plain~\cite{abdelfattah2020zero} & 0.240 & 0.311 & 0.394 &  - \\ \cline{2-7} 
     & \multirow{2}{*}{Architecture-level} & jacob\_cov~\cite{mellor2020neural} & 0.066 & 0.608 & 0.230  & - \\ \cline{3-7} 
     & & relu\_logdet~\cite{mellor2021neural} & 0.290 & 0.611 & 0.539 & - \\ \Xhline{1.5pt}
      \end{tabularx}
  \end{center}\label{analyze}
  \end{table*}

\subsection{Benchmark Datasets}\label{eva_benchmark}
EEMs are often evaluated on public benchmark datasets to perform fair comparisons. Specifically, the benchmark datasets include a lot of architecture-performance pairs in the specific search spaces. The architecture-performance pairs in these benchmark datasets can be used to evaluate the prediction accuracy of EEMs. The commonly used benchmark datasets includes NAS-Bench-101~\cite{ying2019bench}, NAS-Bench-201~\cite{ying2019bench}, NAS-Bench-301~\cite{siems2020bench} for image classification, NAS-Bench-NLP~\cite{klyuchnikov2022bench} for natural language processing, and NAS-Bench-ASR~\cite{mehrotra2020bench} for automatic speech recognition.

These benchmark datasets all use the cell-based search space. Specifically, they stack cells to form the architecture, and only the structure of the cell can be searched. Each cell can be treated as a DAG. Based on the position of the operation in DAG, the cell-based search spaces can be divided into Operation on Node (OON) and Operation on Edge (OOE) search spaces. Specifically, the node is regarded as the operation while the edge is treated as the connection between operations for OON. The edge represents the operation while the node is regarded as the connection for OOE. We summarized the mentioned benchmarks in Table.~\ref{tbl_benchmark}.

\subsection{Comparisons on Benchmark Datasets}\label{eva_comp}
In order to make audiences more intuitively compare the performance of various EEMs, we collect the KTau of them on NAS-Bench-101, NAS-Bench-201, and NAS-Bench-301. Please note the results are collected from the original paper or some papers that report the results of these EEMs. We do not report the results of the downscaled model method, network morphism, gradient-based method, and population memory. For the the downscaled model method, we do not have enough computing resources to re-evaluate the downscaled models in these benchmark datasets. For network morphism, gradient-based method, and population memory, they cannot be used to accelerate all architectures in these benchmark datasets. Specifically, network morphism can only be applied to accelerate the networks morphed from the parent network. The gradient-based method is also a search strategy. It can only be used to accelerate the searched architectures. Performance memory can only be applied to query the performance of architectures that have been evaluated. As a result, the experimental results of these methods are not reported. 

The experimental results are presented in Table.~\ref{analyze}, and we will discuss each-shot method sequentially. (1) \textit{$N$-shot methods:} These methods still need to evaluate all search architectures, which typically consume more time compared to others. Specifically, learning curve extrapolation methods achieve inferior results than downscaled dataset methods. This is because the early-stop learning curve cannot properly reflect the final performance, while CIFAR10 has similar features as ImageNet and is widely used as a proxy pair. In addition, the performance of early stopping at 1/2 is superior to stopping at 1/4. This means more training iterators can match the final learning curve. (2) \textit{Few-shot methods:} The performance predictor needs additional initialization time, but it has the best performance among all methods. Specifically, HOP~\cite{chen2021not} achieves $0.813$ and $0.891$ KTau values, which present the optimal results on NB101 and NB201, respectively. (3) \textit{One-shot methods:} These methods require evaluation time to get the performance of a big supernet, and the path-bathed methods can get similar results as the early stop strategy. (4) \textit{Zero-shot methods:} These methods are regarded as the most time-saving ones because they do not rely on any GPU-based evaluation process. However, this category of methods does not perform as well as other categories. Since they evaluate some statistical metrics, which cannot reflect the real accuracy of the architecture. Note that parameter-level methods even get the negative KTau value on NB101 and NB301.

{The performance of each-shot methods is affected by various factors. Specifically, the similarity between the proxy and original dataset and the representativity of the sub-dataset can highly affect the performance of the downscaled dataset method. In addition, the ability of the performance predictor can be affected by the encoding strategy of the architecture and regression model. Moreover, the path sampling strategy would influence the performance of the path-based methods. Finally, as for the zero-shot method, their performance relies on the designed scoring functions. If the function can precisely reflect the real accuracy of the architecture, the zero-shot methods can acquire great performance.}

\section{Challenges and Future Directions}\label{direction}
In summary, current research efforts are primarily focused on designing EEMs with high prediction accuracy and efficiency on some benchmark datasets like NAS-Bench-101, NAS-Bench-201, and NAS-Bench-301, and have achieved considerable success. Firstly, the performance of neural architectures in existing architecture benchmark datasets is obtained on some small-scale datasets such as CIFAR10. These benchmark datasets cannot reflect the performance of EEMs in real-world scenarios. Secondly, the current EEMs face difficulties in achieving cross-domain prediction of neural architectures across multiple different search spaces. This increases the time complexity because the EEM has to be redesigned for each specific search space. Moreover, the current EEMs lack the capability to evaluate the performance of a neural architecture on multiple datasets within a single search space. This has increased the time required to build EEMs on multiple datasets even for architectures within the same search space. Lastly, the existing EEM methods also struggle with representing multiple types of neural architectures uniformly, which affects their ability to effectively mine valuable architecture data. In this section, we will discuss these challenges in detail.

\subsection{Effectiveness Validation}
The current effectiveness validation for EEMs is not convincing for most researchers. Specifically, the benchmarking datasets are used for effectiveness validation and fair comparisons. However, the existing architecture dataset cannot conform to the practical sceneries because of their small scales. For example, NAS-Bench-101 only involves three types of operations (i.e., convolution 1$\times$1, convolution 3$\times$3, and max pooling). It limits the max number of nodes and edges to seven and nine, respectively. The NAS-Bench-101 search space is so small that it only consists of $423,624$ architectures. NAS-Bench-201 only includes four different operations, and the architecture only has four vertices. The NAS-Bench-201 search space only includes 15,625 architectures, which is even smaller than NAS-Bench-101. In contrast, the search spaces applied in real application scenarios, such as the search space of NASNet, MobileNet, or Transformer, are more complex. They are generally several orders of magnitude more than NAS-Bench-101 and NAS-Bench-201 in quantity. For example, the NASNet search space designs an operation set that contains $15$ operations and five nodes excluding the input and output nodes. The MobileNet search space, as a block-based search space, contains multiple choices for each block. The search space size is about $10^{39}$ with a block size of five. As a result, the results on the existing benchmarking datasets cannot reflect the effect of the EEMs on real scenarios. There exists the need to construct a larger architecture dataset to assist in the validation of the proposed EEMs. 

\subsection{Cross-domain Prediction}
Cross-domain prediction mainly refers to predicting the performance of the architectures in different search spaces. In existing EEMs, only the downscaled dataset method, the downscaled model method and population memory have the ability of cross-domain prediction. This is because these methods do not care about the domains of the architectures. Other EEMs are specific to a target search space and lack cross-domain ability. For example, the designed function-preserving operations in network morphism only can be used in a specific structure. If this structure is not included in the architecture, the network morphism method cannot be used. The performance predictors generally design encoding methods and regression models for specific search spaces. As a result, they cannot be applied or have poor results in other search spaces. The one-shot method can only predict the performance of the subnets contained in the supernet. The cross-domain ability of zero-shot methods also cannot be guaranteed because the measurement of the properties for different types of architectures is different. This means that we have to rebuild an EEM once the search space changes. This is labor-intensive and computationally expensive. There are works that explore the methods of cross-domain prediction. For example, Han \textit{et al.}~\cite{han2021transferable} represented the candidate CNNs as a computation graph that consists of only primitive operators. Then, they proposed a semi-supervised graph representation learning procedure to predict the architectures from multiple families. However, the method is limited to the cell-based search space. How to develop a cross-domain EEM is still a challenging issue.

\subsection{Multi-task Prediction}
Multi-task prediction refers that EEMs require predicting the multiple performance values in multiple tasks of the same architecture. Moreover, there is an inner correlation between the multiple labels of the same architecture. The multi-task situation often appears in multi-task learning. 
Specifically, multi-task learning leverages the useful information between different but related tasks to improve the generalizability of networks~\cite{zhang2021survey}. Multi-task learning has become a hot topic because it can save computational overhead by applying one network to multiple tasks. Furthermore, training one network on multiple tasks can also improve generalization. In multi-task learning, we need to estimate the performance of the architectures on multiple tasks. 
Some researchers have observed this problem and work on it. For example, Huang \textit{et al.}~\cite{huang2022arch} embedded the tasks as a part of the input to achieve the prediction of multiple labels. 
Shala \textit{et al.}~\cite{shala2022transfer} proposed a transferable performance predictor with deep-kernel Gaussian processes to predict the performance architectures on different datasets.
To promote the development of multi-task prediction, Duan \textit{et al.}~\cite{duan2021transnas} proposed a benchmarking dataset (called TransNAS-Bench-101). Specifically, TransNAS-Bench-101 involves the performance of $51,464$ architectures across seven tasks such as classification, regression, pixel-level prediction, and self-supervised tasks. However, the work on this is fairly limited. With the success of multi-task learning in real-world deployment scenarios, we believe that the EEMs for multi-task are a promising future research direction.

\subsection{Uniform Representation}
The performance of the architecture largely depends on the design of the architecture. As a result, how to extract meaningful information from the architectures is critical for the prediction of NAS. To mine the architectures, we first need to provide the representation method to describe the architectures. Although many powerful representation methods have been proposed, they can only represent the architecture in a specific type of search space. For example, the commonly-used adjacency matrix encodings~\cite{white2020study} can only be used to represent the cell-based architectures. This prevents the researchers from learning knowledge of various architectures at the same time. 
Furthermore, this is not conducive to improving the transferability of the EEM. To overcome the problem, Sun \textit{et al.}~\cite{sun2021arctext} designed a unified text method to describe the CNN. Concretely, it designed four units to describe the detailed information of each layer in CNNs. Moreover, it provided a unique order of layers to make the topology information constant. However, this method cannot describe other types of architectures in addition to CNN. The research on the uniform expression for various architectures is still in the early stage for the field of NAS.

\section{Conclusion}\label{conclusion}

This paper gives a comprehensive survey of EEMs. Specifically, based on the number of architectures trained, we categorize the EEMs into the $N$-shot methods, few-shot methods, one-shot methods, and zero-shot methods. The $N$-shot methods require training every searched architecture and mainly consist of the downscaled dataset methods, downscaled model methods, learning curve extrapolation methods, and network morphisms. The few-shot methods only need to train a smaller number of architectures than TEM. It mainly includes the performance predictor and the population memory. The one-shot methods merely require training one architecture, and mainly include the path-based methods and the gradient methods. The zero-shot methods involve no training, thus further reducing the runtime. They can be divided into the parameter-level method and architecture-level method. Furthermore, we review the evaluation metrics and benchmark datasets for EEMs, and report the results on these benchmark datasets to intuitively show the performance of various EEMs. Lastly, we summarize the challenges and future research directions of the existing EEMs. Specifically, the challenges mainly include effectiveness validation, cross-domain prediction, multi-task prediction, and uniform representation.

\ifCLASSOPTIONcaptionsoff
  \newpage
\fi

\bibliographystyle{IEEEtran}
\bibliography{reference}

\end{document}